\UseRawInputEncoding
\documentclass[oneside,12pt]{Classes/RoboticsLaTeX}

\usepackage{amsmath}
\usepackage{amsfonts}
\usepackage{algorithm}
\usepackage{algorithmic}
\usepackage{graphicx}
\usepackage{hyperref}
\usepackage{subfig}
\usepackage[numbers]{natbib}

\ifpdf
    \pdfcatalog {/PageMode (/UseOutlines)
                 /OpenAction (fitbh) }
\fi

\title{Obstacle Detection and Tracking in Marine Environments}

\ifpdf
  \author{\href{mailto:yara.ala96@gmail.com}{Yara AlaaEldin}\\ \href{mailto:Enrico.Simetti@unige.it}{Enrico Simetti}, \href{mailto:francesca.odone@unige.it}{Francesca Odone}\\
  }
  \collegeordept{\href{http://www.dibris.unige.it}{DIBRIS - Department of Computer Science, Bioengineering, Robotics and System Engineering}}
  \university{\href{http://www.unige.it}{University of Genova, Italy}}
  \crest{\includegraphics[width=30mm]{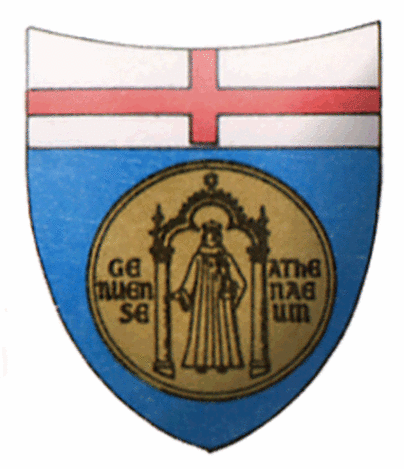}}
\else
  \author{Yara AlaaEldin\\ Enrico Simetti, Francesca Odone}
  \collegeordept{DIBRIS - Department of Computer Science, Bioengineering, Robotics and System Engineering}
  \university{University of Genova}
  
  \crest{\includegraphics[width=30mm]{Figures/System/logo_unige.png}}
\fi

\degree{Masters of Science in Robotics Engineering}
\degreedate{October 2022}

\begin{document}


\maketitle

\newpage\null\thispagestyle{empty}\newpage

\setcounter{secnumdepth}{3}
\setcounter{tocdepth}{3}

\frontmatter


\begin{acknowledgements}

In the name of Allah the merciful,

Firstly, I thank my parents for everything they did for me in order to be a well-educated and influential person. You are the indirect reason behind everything I achieve. 

I would like to express my thanks and appreciation to my supervisors, professor Simetti and professor Odone, for their continuous guidance and stimulating suggestions in every part of the project. I would like also to thank Issa Mouawad for his help and support in this project from the beginning.

Big thanks are due to the professors who taught us during this masters program. This work is a result of the accumulative experience gained from all the courses taught in this challenging program.

I must express my gratitude to my brothers, sister, family, and friends in Italy and in Egypt for providing me with unfailing support
and continuous encouragement throughout the two years of study and
through the process of working on this project. This accomplishment
would not have been possible without them.

In the end, I am grateful to God for the good health and well-being that were necessary to complete this work.

\end{acknowledgements}


\begin{dedication}

This work is dedicated to my parents for their endless love and support.
\end{dedication}





\begin{abstracts}

Developing a robust and effective obstacle detection and tracking system for Unmanned Surface Vehicle (USV) at marine environments is a challenging task. Research efforts have been made in this area during the past years by GRAAL lab at the university of Genova that resulted in a methodology for detecting and tracking obstacles on the image plane and, then, locating them in the 3D LiDAR point cloud. In this work, we continue on the developed system by, firstly, evaluating its performance on recently published marine datasets. Then, we integrate the different blocks of the system on ROS platform where we could test it in real-time on synchronized LiDAR and camera data collected in various marine conditions available in the MIT marine datasets. We present a thorough experimental analysis of the results obtained using two approaches; one that uses sensor fusion between the camera and LiDAR to detect and track the obstacles and the other uses only the LiDAR point cloud for the detection and tracking. In the end, we propose a hybrid approach that merges the advantages of both approaches to build an informative obstacles map of the surrounding environment to the USV.

\end{abstracts}

\tableofcontents
\listoffigures

\mainmatter

\chapter{Introduction}
\label{chap:introduction}
\ifpdf
    \graphicspath{{Introduction/Figures/PNG/}{Introduction/Figures/PDF/}{Introduction/Figures}}
\else
    \graphicspath{{Introduction/Figures/EPS/}{Introduction/Figures}}
\fi

Ranging from scientific exploration to search-and-rescue operations and national security missions, the applications that require robust and reliable autonomous Unmanned Surface Vehicles (USV) are increasing every day. Such vehicles can replace humans in performing tedious and risky tasks at sea while at the same time reduce the error probability. 

The autonomous navigation system of such vehicles is an essential and common block that has to be perform efficiently regardless of the application. The most basic role of such navigation system is to detect the surrounding obstacles, detect their classes (e.g. boat, ship, sailboat, buoy, etc), identify them with tracking IDs, estimate their heading trajectories, build a scene map of the surrounding environment containing obstacles locations, boundaries, and headings, plan a feasible path for the USV towards goals given the scene map, and follow the planned path towards the goal using control modalities. 

To achieve such basic yet complex mission different on-board sensors (e.g. RGB camera, IR camera, depth camera, stereo camera, 2D LIDAR, 3D LIDAR, RADAR, etc) can be mounted on the USV and used for environment perception. The data coming from such sensors can be processed and fused using several techniques achieving varied results in scene understanding and obstacles avoidance. This is currently a hot area in research for which new papers are being published proposing methodologies for obstacles detection and tracking in maritime environments.

In the previous years, a research team from the university of Genova developed a method~\cite{sorial2019towards} for the detection and tracking of obstacles in marine environments using only two on-board sensors: an RGB camera and a 3D LIDAR. The method relies on state- of-the-art object detection CNN (Convolutional Neural Network) YOLO~\cite{redmon2018yolov3} to detect obstacles in the image plane and an obstacle tracking procedure based on both distance and appearance to track the detected obstacle across video frames. After that, transformation from camera frame to 3D frame is done to locate the point cloud clusters of the detected obstacles and their 3D bounding boxes. Finally, a scene map with the obstacles bounding boxes and their headings is produced that can be the input of a path planning algorithm. 

The method has been validated and evaluated qualitatively and quantitatively using well-established metrics on manually collected data. Nevertheless, the system is still in its development state and needs more strict evaluation on various marine conditions especially given the recently published datasets like the MIT marine dataset~\cite{MIT} that contain synchronized video and lidar data collected in various marine conditions. Such datasets can be used for benchmarking our method. Furthermore, ideas of enhancement on the method can be proposed and implemented drawing inspirations from the recent research advancements in the field.

\section{Problem Statement}

\begin{figure*}%
    \centering
    {\includegraphics[scale=0.5]{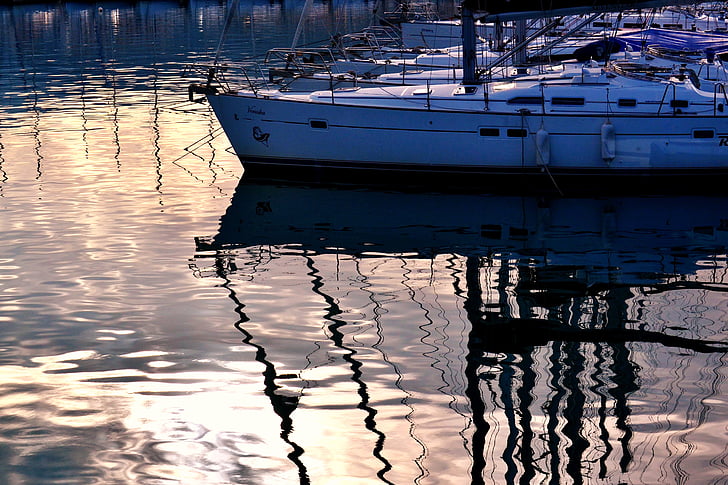}}
    \caption{Water reflections}%
    \label{fig:reflec}%
\end{figure*}

\begin{figure}%
    \centering
    \subfloat{{\includegraphics[width=6cm, height=4cm]{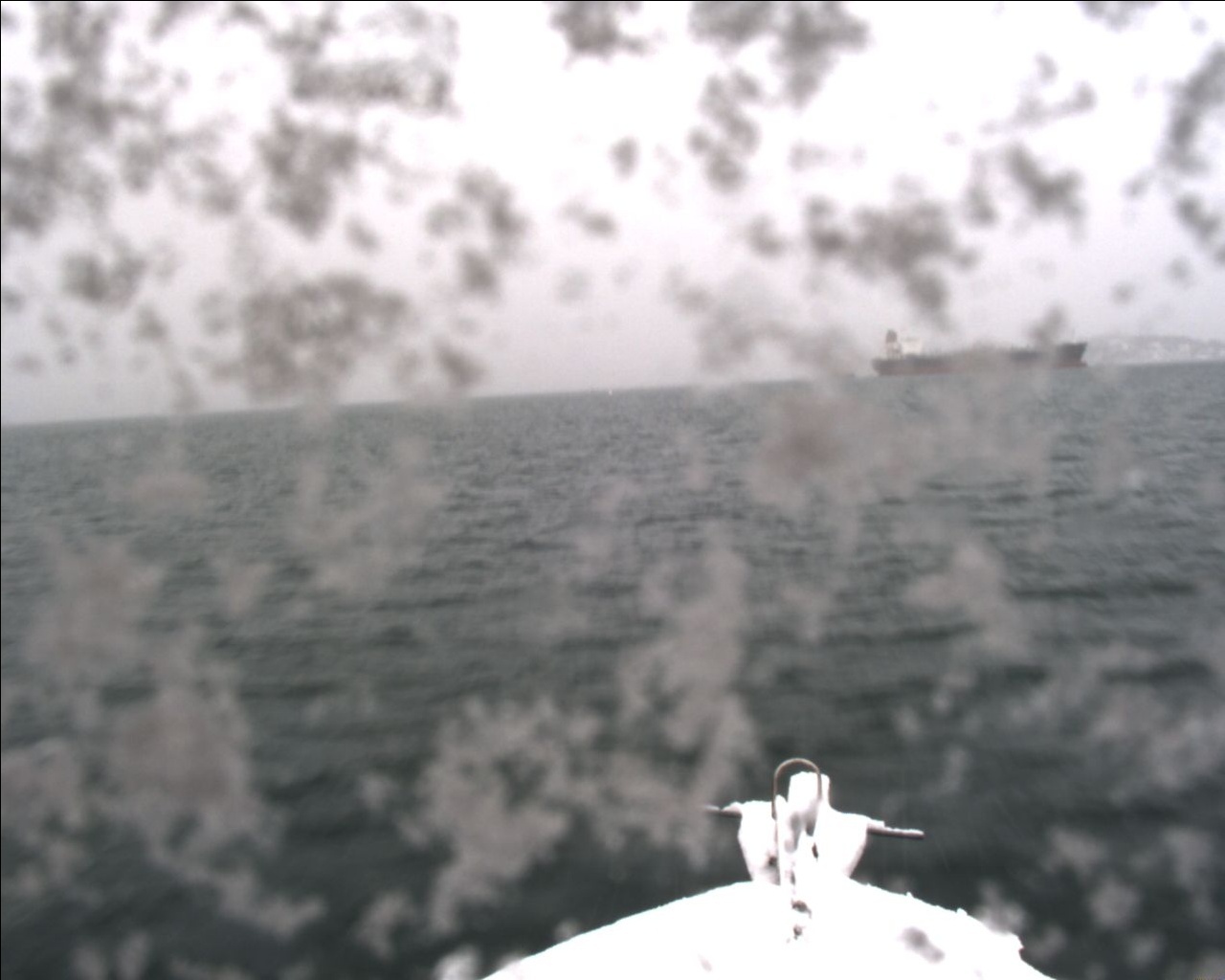} }}%
    \qquad
    \subfloat{{\includegraphics[width=6cm, height=4cm]{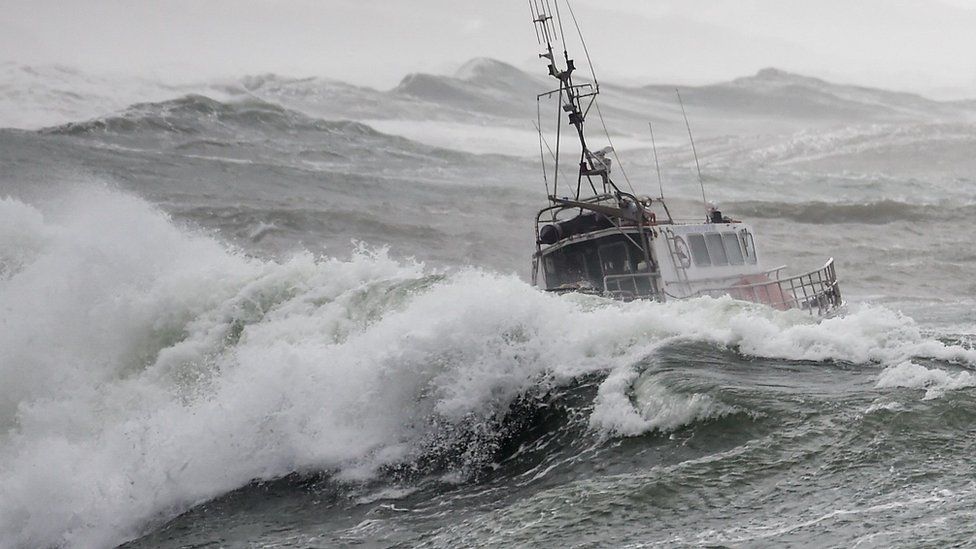} }}%
    \caption{Extreme weather conditions in the sea}%
    \label{fig:weather}%
\end{figure}

\begin{figure}%
    \centering
    \subfloat{{\includegraphics[width=6cm, height=5cm]{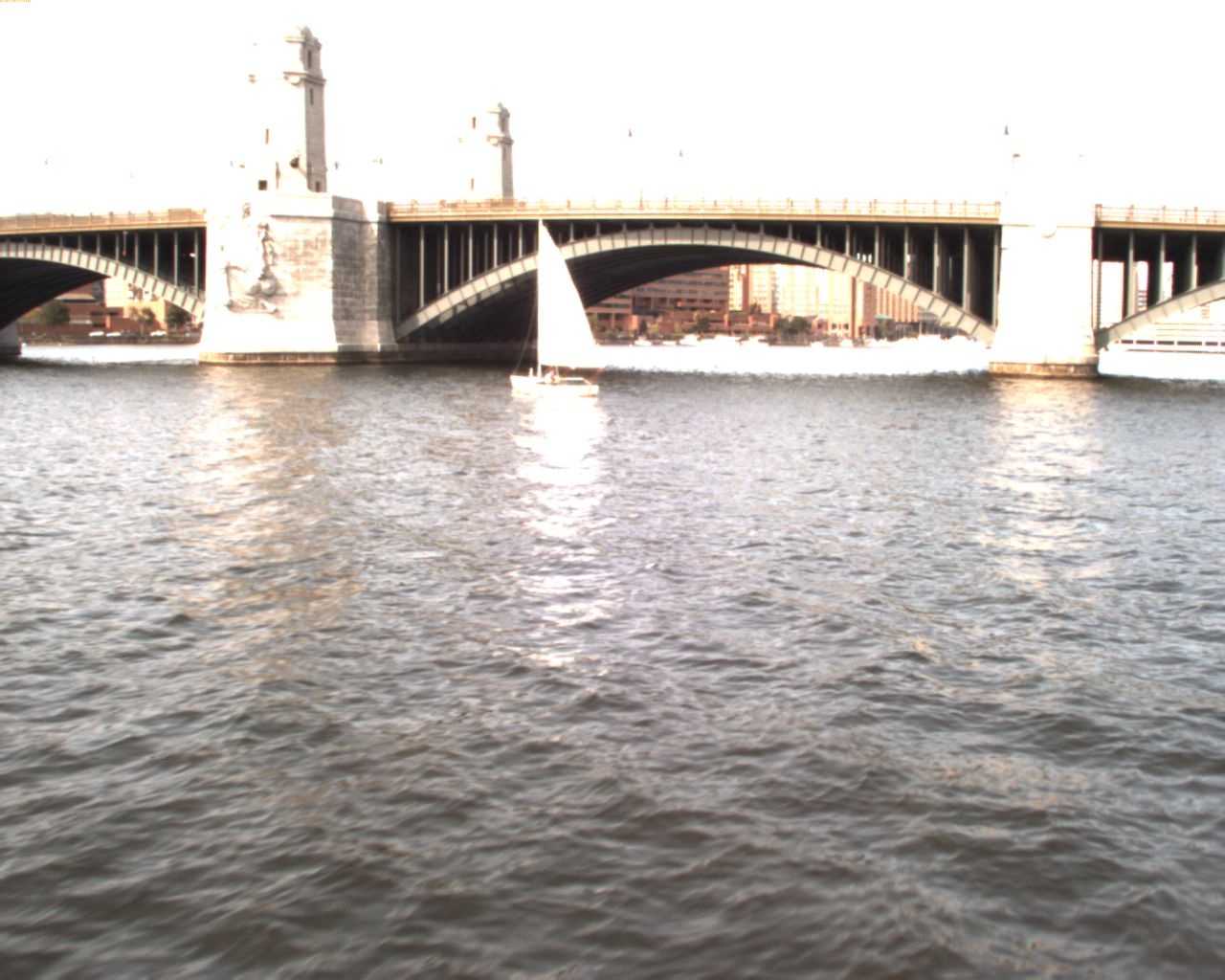} }}%
    \qquad
    \subfloat{{\includegraphics[width=6cm, height=5cm]{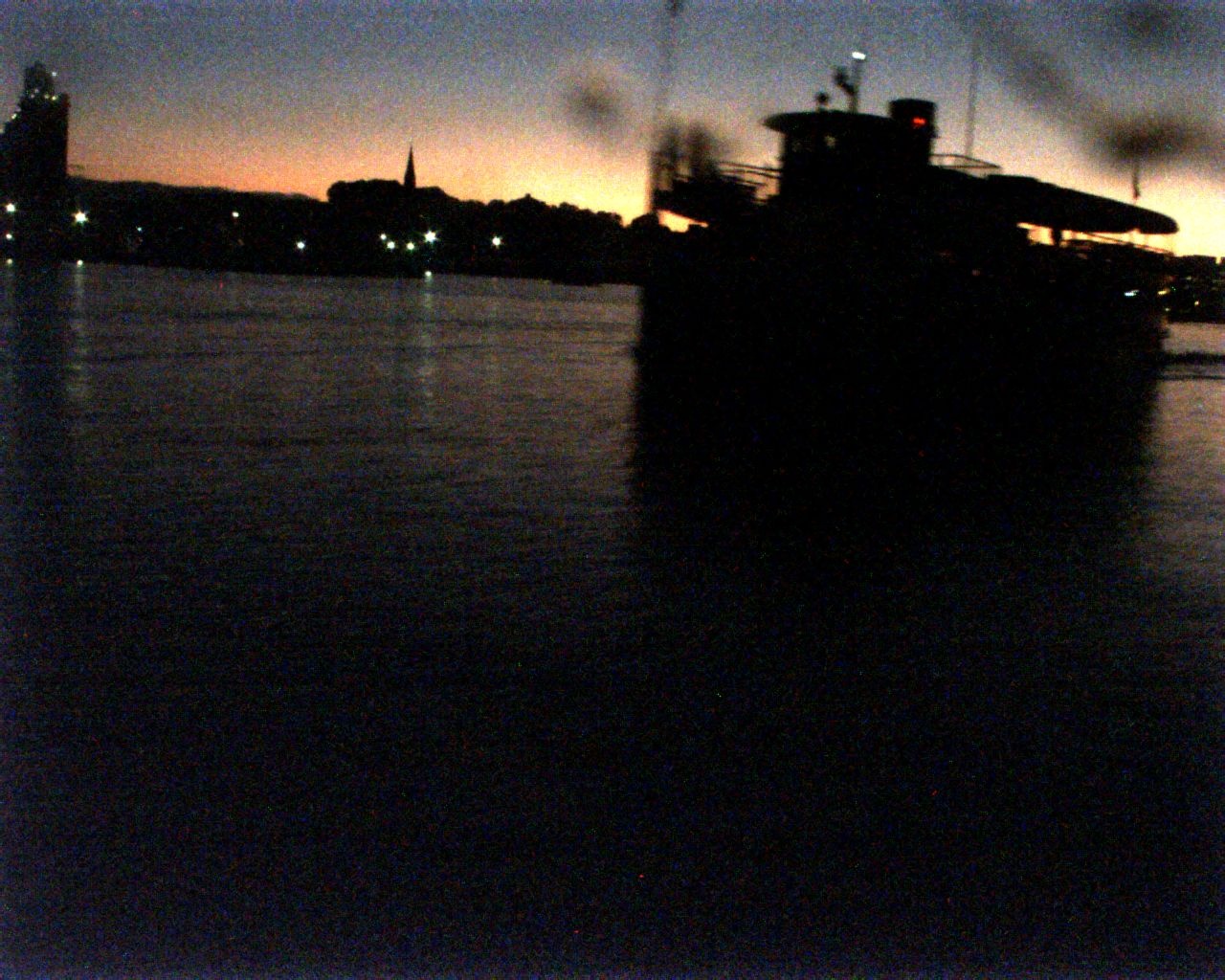} }}%
    \caption{Various lighting conditions in the sea (a) sea glare (b) sunset}%
    \label{fig:light}%
\end{figure}

\begin{figure}%
    \centering
    {\includegraphics[width=7.5cm, height=6cm]{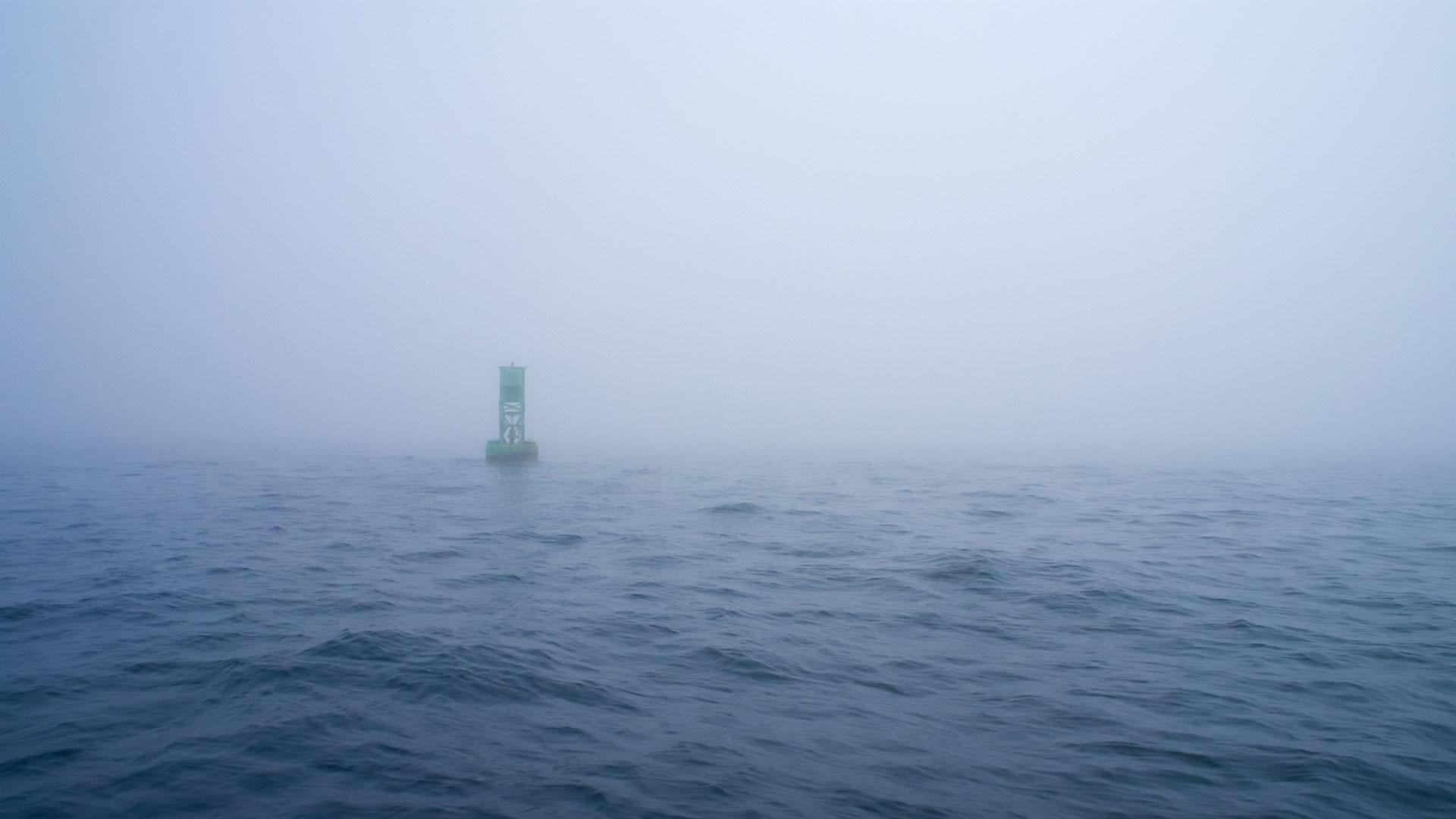} }
    \caption{Sea Fog}%
    \label{fig:fog}%
\end{figure}

Autonomously navigating USV at sea environments is a non straightforward task. Differently from urban environments, the maritime environments have unique set of challenges. Environmental disturbances like sea fog, rain, wind, waves and currents greatly affect the visual perception of the USV(see figures~\ref{fig:fog} and ~\ref{fig:weather}). Fog and rain cause unclear vision for the camera that may result in missing the detection of some obstacles. Wind, waves, and currents lead to non-stable and divergent movements of the USV and this poses a big challenge to the obstacles tracking module. Various lighting conditions, that change depending on the weather and time of the day(see figure~\ref{fig:light}), can cause problems for the obstacles detection module. In addition, water reflections of objects on the sea surface can deceive the detection module and cause false detections (see figure~\ref{fig:reflec}). 

Using range sensors like LiDAR for perception also has its own set of challenges. The LiDAR point clouds received from maritime environments suffer from sparsity due to the large separation distances between the USV and the objects at sea. These sparse point clouds under-represent the point cloud clusters of obstacles and this may lead to non-accurate estimation of the obstacles bounding boxes or missing the detection of small obstacles. 

Furthermore, the USV is required to operate and react to obstacles in real time. This puts a restriction on the time complexity of the used algorithms for obstacles detection and tracking. In many cases, the optimal methods for obstacles detection and tracking (e.g. detection using image semantic segmentation) require long processing times compared to less accurate methods. The hardware capabilities pose limits as well on the computational requirements allowed for detection and tracking algorithms.

\section{Current System State}

To overcome the drawbacks of both visual and range sensors, the developed detection and tracking procedure fuses the data coming from both types of sensors (visual: video camera, range: 3D LiDAR) to obtain a powerful scene understanding. Obstacles are detected in the video frames using trained state-of-the-art object detection CNN YOLO. The detector extracts the obstacles of class "boat" and detect their bounding boxes in each video frame. The object tracking module gets the obstacles detections and assign tracking IDs to the obstacles given the distance and appearance similarity with the history of tracks. The module implements a Kalman Filter prediction for the missing tracks to account for missing detection of obstacles in some frames. The output of the tracking module is projected to the 3D reference frame and the point cloud clusters corresponding to each detected obstacle are extracted and enclosed with 3D bounding boxes. The scene map is generated with the obstacles bounding boxes and their headings that can be the input to a path planning module.

\begin{figure}%
    \centering
    {\includegraphics[width=7.5cm, height=6.5cm]{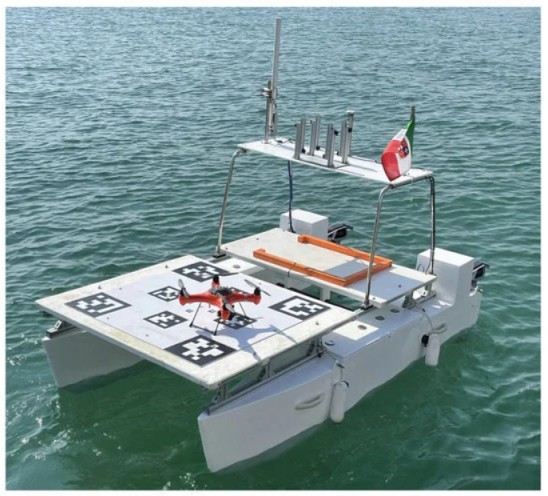} }
    \caption{ULISSE Catamaran}%
    \label{fig:ulisse}%
\end{figure}

The method has been validated and tested on recorded data by a USV (codenamed ULISSE) (see figure ~\ref{fig:ulisse}) using ROS middleware. ULISSE was developed by Genoa Robotics And Automation Laboratory (GRAAL)~\footnote{\url{https://graal.dibris.unige.it/}} in collaboration with other research, industry and government partners. The recorded data included HD videos (at 1032x778 pixels) at a maximum frame-rate of 31fps synchronized with lidar data that covers 360 degrees horizontally with 30 degrees vertical coverage. The data was acquired near harbour on a single day and represented different obstacles and approaching angles. After collection, the data was manually annotated for evaluation.

The video tracking module was evaluated quantitatively on the collected data using two well-known metrics: Average Precision (AP), and Multi-Object Tracking (MOT)~\cite{leal2015motchallenge}. The tracking method showed better results as compared to bare YOLO detector and SORT tracking method~\cite{wojke2017simple}. The 3D localization module, on the other hand, was not strictly evaluated due to the limited amount of available lidar data. 

In addition, execution time analysis was performed on each module separately (detection module, tracking module, 3D localization module). It was noted that the time required by the tracking module depends on the number of objects within a frame.

Currently, each module is implemented individually and runs off-line given the output of the previous module. The detector YOLOv3 is implemented in C language and takes as input the collected video frames. The tracker is implemented on Python 3.6 and takes as input the output detections from YOLO. The 3D localization is implemented on MATLAB and takes as input the output tracks of the tracking module.

\section{Contribution}

In this work, our mission can be stated as the following: 
\begin{itemize}
    \item investigate and analyse the recent research work done in the field including the types of sensors used, the processing techniques of the sensors data, the hardware specifications, the algorithms for detection and tracking of obstacles and the sensor fusion techniques. 
    \item explore the available public marine datasets that can be used for the evaluation of our system especially those that contain synchronized lidar and video data. Use this data to benchmark our system to compare its performance with the other methods evaluated on the same data. 
    \item implement and integrate the system blocks (detection, tracking, and 3D localization) on ROS platform where the output of each module is passed to the following one and the system can be tested in real-time. 
    \item propose ideas of enhancements on the system after investigation of the evaluation results and test the effectiveness of these ideas. 
    
\end{itemize}

The remainder of this thesis is organized as follows:
\begin{itemize}
    \item In \textbf{Chapter 2}, a literature review of the state-of-the-art research in the field is presented and analyzed.
    \item In \textbf{Chapter 3}, the previously developed obstacles detection and tracking pipeline is explained in details.
    \item In \textbf{Chapter 4}, we present the software architecture of the developed system on ROS.
    \item In \textbf{Chapter 5}, the evaluation results of the system on various datasets are introduced. In addition, we present a thorough experimental analysis of the results obtained using two approaches; one that uses sensor fusion between the camera and LiDAR to detect and track the obstacles and the other uses only the LiDAR point cloud for the detection and tracking.
    \item In \textbf{Chapter 6}, the thesis is concluded and promising directions for future efforts are provided
\end{itemize}

\chapter{State of The Art}
\label{chap:first}
\ifpdf
    \graphicspath{{Chapter1/Figures/PNG/}{Chapter1/Figures/PDF/}{Chapter1/Figures/}}
\else
    \graphicspath{{Chapter1/Figures/EPS/}{Chapter1/Figures/}}
\fi

In this chapter, a review of research papers in the field of Unmanned Surface Vehicle (USV) obstacle detection and tracking in maritime environments is presented. The review focuses on recently published research papers in that field. Overall, semantic segmentation and horizon detection methods are widely used in research for detecting the water region in images, and consequently finding obstacles that are inside in the water region under the horizon line. However, this method suffers from many problems including: long inference time because the semantic segmentation neural network is relatively more complex than the usual bounding-box-based object detection networks, vehicle unstability on the water leading to horizon detection problems, errors in horizon detection when the vehicle is near the shore, different weather conditions leading to unclear water boundaries, and mirroring on the sea surface leading to errors in water region estimation. Moreover, semantic segmentation based methods don't predict the obstacle class and they have low ability in small obstacles detection.

Although RADAR and LiDAR are heavy and power consuming sensors, the clustering of their point clouds was also used as a method for obstacles detection at sea. The main issue with LiDAR is the sparsity of the point cloud perceived from the marine environment that makes it difficult to tune the clustering parameters. Also, the clustering algorithms generally suffer from bad performance in cluttered and near-shore environments. They also require a lot of tuning for parameters like the max distance between two points in the same cluster and the minimum distance between two clusters, and other parameters. However, the main advantage of this method with respect to camera-based methods is that the LiDAR and RADAR point clouds are unaffected by weather conditions. 

The main reason why bounding-box-based object detection deep learning methods were not extensively investigated in the literature is the lack of big-sized and appropriately annotated marine datasets for obstacles detection. The abundance of appropriately annotated datasets is the reason why there are a lot of research advances in the field of object detection in unmanned ground vehicles. This is not the case for marine vehicles where there is no availability of such big-size datasets that cover all the different complex situations and environmental conditions in the sea. To overcome this challenge, some researches used already trained object detection neural networks like YOLO~\cite{redmon2018yolov3,bochkovskiy2020yolov4} and Faster R-CNN~\cite{ren2015faster} trained on common object detection datasets like MS-COCO ~\cite{lin2014microsoft}, and fine-tuned these models on small collected and annotated datasets of marine obstacle detection. Recently, multiple marine datasets were released and they can be used for properly training (or fine-tuning) deep learning neural networks for obstacles detection at sea. 

In the next section, brief summaries of the marine object detection and tracking methods used in the reviewed papers are presented with a comparison table including the main features of each method. Next, the publicly available marine datasets are presented along with their descriptions. Finally, some benchmarking results are presented that show the performance of different methods on some of the presented datasets.

\section{Reviewed Work}
\begin{figure}%
    \centering
    {\includegraphics[width=8cm, height=19cm]{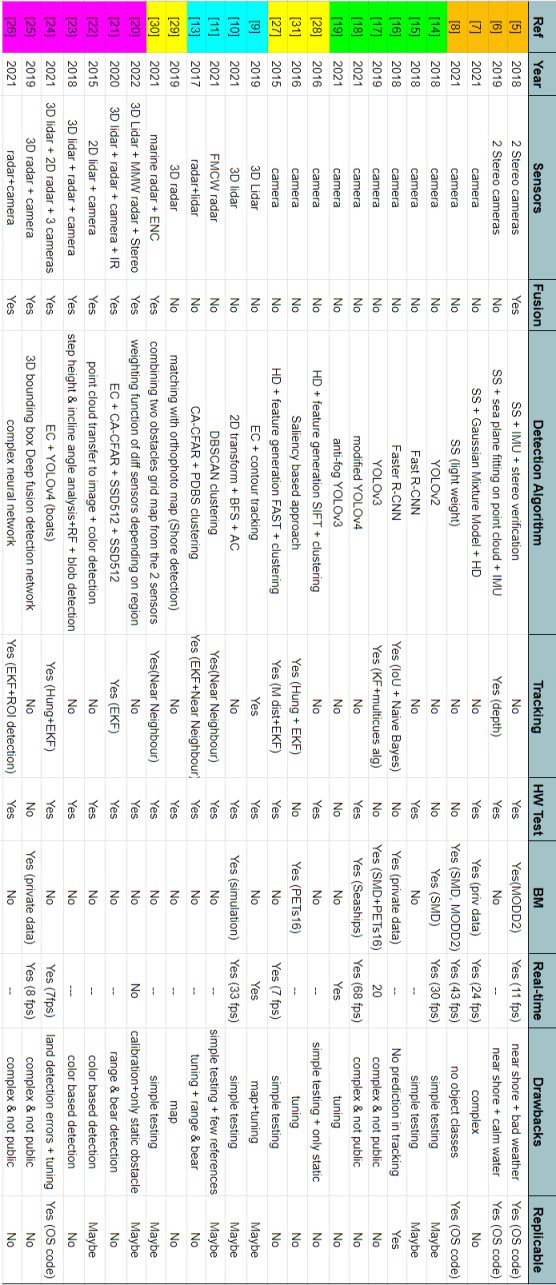} }
    \caption{Reviewed Papers Comparison Table}%
    \label{fig:table}%
\end{figure}

\begin{figure}%
    \centering
    {\includegraphics[width=7.5cm, height=6.5cm]{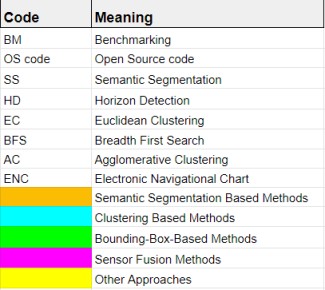} }
    \caption{Key Table for table~\ref{fig:table}}%
    \label{fig:keytable}%
\end{figure}

In this section I summarize the relevant reviewed papers that are recently published in the field of marine obstacles detection and tracking. An overall comparison between the investigated papers is presented in table~\ref{fig:table}. Table~\ref{fig:keytable} shows the meanings of abbreviations and color codes used in table~\ref{fig:table}.

\subsection{Semantic Segmentation}
\begin{figure}%
    \centering
    {\includegraphics[width=7.5cm, height=6.5cm]{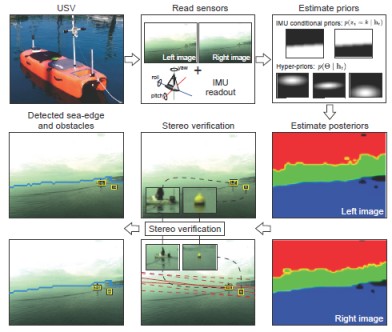} }
    \caption{Detection Method used in~\cite{bovcon2018stereo}}%
    \label{fig:fig1_1}%
\end{figure}

\begin{figure}%
    \centering
    {\includegraphics[width=7.5cm, height=10cm]{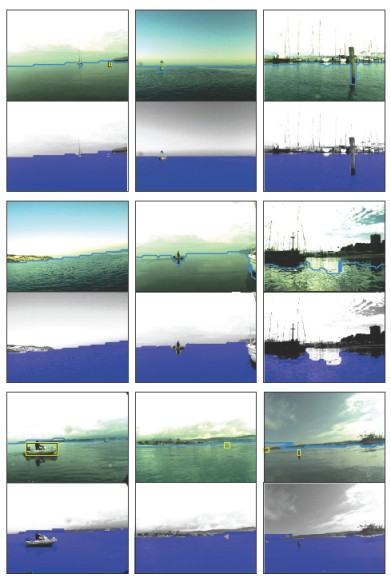} }
    \caption{Evaluation Results in~\cite{bovcon2018stereo}}%
    \label{fig:fig1_2}%
\end{figure}

\textbf{Semantic segmentation and horizon detection} methods were adopted in \cite{bovcon2018stereo,muhovivc2019obstacle,liu2021real,chen2021wodis} for obstacles detection at sea. In \cite{bovcon2018stereo}, they used two stereo cameras and an IMU. The IMU reading is used for horizon line estimation in the received camera images. Then, semantic segmentation is applied on the image to extract the water region and the objects lying inside the water (below the horizon line) are considered as obstacles. The detected obstacles are verified by comparing it with the point cloud coming from the stereo cameras, see figure~\ref{fig:fig1_1}. Their method suffers from a lot of False Positive (FP) and False Negative (FN) detections especially near the shore or in bad weather conditions, see figure~\ref{fig:fig1_2}. 

They continue their work in \cite{muhovivc2019obstacle} where they introduce an obstacle tracking approach based on depth signature in the 3D point cloud coming from the stereo cameras, and they discard the non-consistent detections over time. This helped in reducing the FP rate. They also used the stereo camera point cloud to improve the sea plane estimation (horizon line detection) given that the stereo camera detects many points from the sea texture. However, this method faced problems in calm water where the point cloud is very sparce. Also, there were problems in sea plane estimation near the shore, with water reflections, and with big obstacles. Their code is publicly available: \footnote{\url{https://www.vicos.si/resources/modd/}}

In \cite{liu2021real}, they use only one camera as a sensor. They presented a new horizon line estimation based on semantic segmentation. The method assumes a Gaussian Mixture Model for fitting the semantic structure of images coming from the camera and generate a water segmentation mask. The horizon line is estimated from the boundaries of the water region. Obstacles are detected as the objects below the horizon line. They claim that their method outperforms the other horizon line detection methods. Although their method achieved a better real time performance (~24 frame per second), their model is complex and hard to be adopted and it still suffers from the previously mentioned semantic segmentation problems (water reflections, near the shore, and big obstacles problems).

In \cite{chen2021wodis}, they proposed a novel lightweight semantic segmentation neural network named WODIS. This network produces better results than the current state-of-the-art especially with different environment conditions and it is evaluated on different datasets. It produces high accuracy compared to other semantic segmentation methods. However, their method still suffers from near the shore detection problems.
They claim real time capability of their system with 43 frames per second for only detections (no tracking). Their code is publicly available: \footnote{\url{https://github.com/rechardchen123/ASV_Image_Segmentation}}

\subsection{Clustering}
\textbf{Clustering} based detection techniques were adopted in~\cite{wang2019roboat,jeong2021efficient,im2021object}. The main idea is to process the point cloud received from 3D LiDAR or from RADAR and cluster every group of points representing an obstacle in one cluster. In ~\cite{wang2019roboat}, they use Euclidean Clustering algorithm using the library PCL~\cite{rusu20113d}. They are focused on the urban environments. Their method receives the point cloud from 3D LiDAR and filter out the urban sideways using a pre-acquired map of the environment. The obstacle clusters contours are assigned unique IDs that are used for tracking them in successive frames. The tracking is achieved using distance threshold between clusters in the frames and they use Kalman Filter for prediction. The method suffers from the mentioned clustering based problems (a lot of parameters to be tuned) and also it requires a map of the environment. 

\begin{figure}%
    \centering
    {\includegraphics[width=7.5cm, height=6cm]{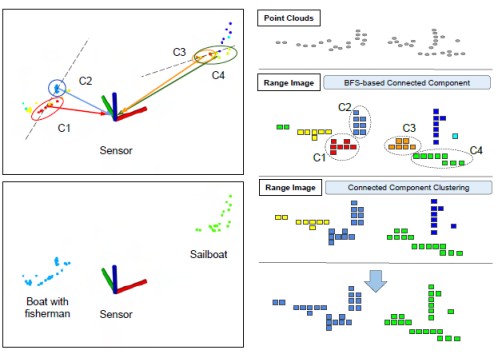} }
    \caption{Clustering Method used in~\cite{jeong2021efficient}}%
    \label{fig:fig2_1}%
\end{figure}

In~\cite{jeong2021efficient}, they detect obstacles by using agglomerative clustering and breadth first search on the LiDAR 3D point cloud after transforming it to 2D point cloud using a spherical projection, see figure ~\ref{fig:fig2_1}. Their method’s evaluation showed better accuracy than Euclidean Clustering with reduced inference time (30 mS per frame). However, they only tested their system on simple scenarios. 

In~\cite{im2021object}, They use FMCW radar for obstacles detection using the algorithm DBSCAN which is an algorithm based on the clustering approach. The clustering algorithm also considers the radar energy level of each point. The tracking of clusters is implemented using the nearest neighbor algorithm. They claim that FMCW radar overcomes the drawbacks of the marine radar that include low update rate and local blind zones.

In~\cite{han2017persistent}, they get the bearing and range of obstacles using both radar and lidar. Lidar is used to detect the obstacles in the shadow zone of the radar.
For radar, the object detection algorithm used is cell averaging constant false alarm rate CA-CFAR after excluding the land area from the radar image. For lidar, PDBS clustering is used for object detection.
They also perform obstacle tracking using Extended Kalman Filter (EKF) for prediction and the global nearest neighbor approach (GNN) for data correlation. The method suffers from the mentioned clustering based problems (a lot of parameters to be tuned) and also it requires a map of the environment.

\subsection{Bounding-Box Detectors}
\begin{figure}%
    \centering
    {\includegraphics[width=8cm, height=5cm]{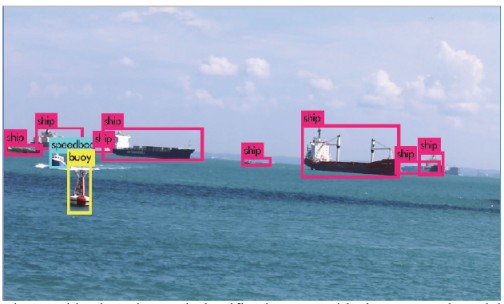} }
    \caption{Sample Result of Fine-tuned YOLOv2 on SMD used in~\cite{lee2018image}}%
    \label{fig:fig3_1}%
\end{figure}

\textbf{Bounding-Box-Based Deep Learning Detectors} were implemented in many papers ~\cite{lee2018image,chen2018development,kim2018probabilistic,qiao2019m3c,liu2021sea,zhang2021sea}. However, the main concern in such papers is the simple evaluation of the trained models. They tested their trained models on relatively simple cases and few of them conducted quantitative evaluation and real-time performance evaluation. In~\cite{lee2018image}, They use YOLOv2 for object detection. They stated that they found it is faster than R-CNN series and gives higher accuracy (no quantitative evaluation was presented in the paper). They fine-tuned the already trained YOLOv2 model on Pascal VOC with Singapore Maritime Dataset (SMD), see figure~\ref{fig:fig3_1}. They tested the detection accuracy of their fine-tuned model on the validation data of SMD. The recall achieved was 0.77 and the real time performance was 30 frames per second. 

In ~\cite{chen2018development}, they built their own hardware USV system and explained its control model in detail. They used camera as the sensor for object detection. Their object detection approach is using fast R-CNN trained on Pascal-VOC dataset without fine-tuning. They tested their trained model on SMD dataset, but they did not provide any quantitative evaluation results.

In ~\cite{liu2021sea}, they proposed a modified version of YOLOv4 with Reverse Depthwise Separable Convolution (RDSC) applied to the backbone network of YOLOv4. The complexity of their modified model is simpler than YOLOv4 and it achieved increased accuracy (mAP) of 1.78\% over YOLOv4 and higher detection speed by more than 20\%. The drawback of such method is the difficulty of reproducing their proposed network.

\begin{figure}%
    \centering
    {\includegraphics[width=3cm, height=8cm]{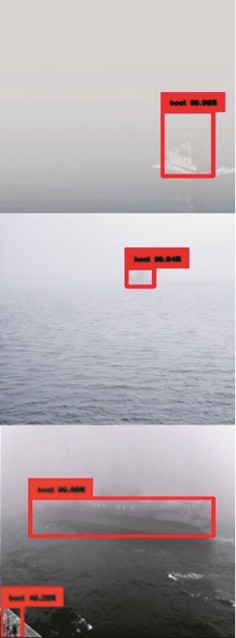} }
    \caption{Sample Result of anti-fog YOLOv3 used in~\cite{lee2018image} on images from internet}%
    \label{fig:figfog}%
\end{figure}

In~\cite{zhang2021sea}, they tackled the problem of object detection under foggy conditions. They generated training samples of foggy images from normal sea images from MS-COCO dataset. Then, they trained a YOLOv3 neural network with the generated samples plus the original MS-COCO dataset . To generate the foggy images, they used the Atmospheric Single Scattering Model. This model has different parameters that must be tuned and they can change from one environment to the other and they depend on the camera installation. Their method requires a dehThey stated that YOLOv3 doesn’t perform fast in the case of multiple targets detection. That's why they used it just for single obstacle detection. They validated their method on test data from the internet, see figure~\ref{fig:figfog}.

Some papers also tackled the problem of obstacle tracking in maritime environments. In ~\cite{kim2018probabilistic}, they built a deep learning system that detects ships using Faster RCNN and proposed a matching algorithm to match bounding boxes between consecutive frames based on IoU. After that, the prediction probability and bounding box class at each frame is determined by combining the confidence probabilities of the bounding boxes in all the previous frames using Bayesian fusion. However, they assume that the ships are moving very slowly so they don’t use any prediction algorithm in their tracking module. In case there is a missing detection in a frame (IoU < threshold), the bounding box of this detection is kept as in the previous frame but after adding an offset to account for the closest matching recently detected bounding box. 
They collected their own sequence dataset from Google images and used it for both training and testing. The testing showed great results in overcoming the mis-classifications of the faster RCNN in some intervals in the video sequence thanks to the tracking module.

In ~\cite{qiao2019m3c}, they propose a joint detection and tracking system for maritime vessels. Their detector is YOLOv3. For tracking, they predict the objects’ locations using one of three motion models: constant velocity, constant acceleration, and curvilinear motion. They formulated different state vectors and transition matrices for each motion model. Given the state vectors at successive frames, a GRU-attention model was trained to output the class probability of each motion model to decide the type of motion of the obstacles. Kalman filter was used for state prediction given the motion model.
They proposed an association model between the detected vessels and the tracks based on multi cues (both long-term cues and short-term cues). 
They tested their system on SMD and PETs2016 datasets and achieved state of the art results compared to other evaluated methods. 
The main drawback of this method is that their models are complex and non reproducible.

\subsection{Sensor Fusion Based Methods}
Many researches ~\cite{liu2022new, han2020autonomous, lebbad2015bayesian, stanislas2018multimodal, clunie2021development,wu20193d,yu2021object} fused data coming from different sensors for increasing both the accuracy and range of detection. This helps also in overcoming the drawbacks of each sensor and taking advantage of each sensor strength points. 

In~\cite{liu2022new}, 3 sensors were used: 3D lidar, MMW radio frequency, and stereo camera. They built a 2D grid representation of the area around the USV and assigned a weighting function for the signal received from each sensor depending on the region, see figure ~\ref{fig:fig15_1}. The weighting function takes into consideration that some sensors are strong in some regions and have blind zones in other regions. However, they only tested their method on static obstacles. A big drawback of such method is that it requires precise calibration between the three sensors to produce the obstacles 2D grid map.

\begin{figure}%
    \centering
    {\includegraphics[width=6cm, height=6cm]{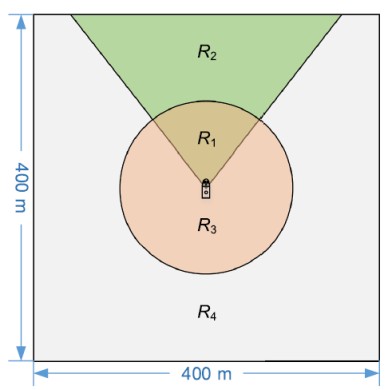} }
    \caption{Perception area schematic diagram of USV obstacle sensors used in~\cite{liu2022new} (R1: perceived by all sensors, R2: perceived by MMW radar and stereo vision, and R3: perceived by 3D lidar)}%
    \label{fig:fig11_1}%
\end{figure}

In~\cite{han2020autonomous}, they use 4 types of sensors: radar, 3D LiDAR, optical camera, and IR camera.  Each one of them has its own object detection algorithm, from which the relative bearing and range of the detected obstacle are extracted. All bearing and range data coming from the sensors are used for EKF based tracking filter for trajectory estimation of the obstacle. Then, the estimated states of all the tracking filters are fused to get a fused estimated state of the detected obstacle. This estimated state enter as input to the collision avoidance module for adjusting the USV navigation.
For radar, the object detection algorithm used is cell averaging constant false alarm rate CA-CFAR after excluding the land area from the radar image.
For LiDAR, the object detection algorithm used is point-distance-based-segmentation PDBS. For both optical and IR cameras, deep learning neural network is used (SSD512). They claimed that SSD network is more effective than YOLO and faster R-CNN, but they didn't provide any quantitative evaluation to validate that. They made some field experiments but no benchmarking results were presented.

In~\cite{lebbad2015bayesian}, they used camera and 2D LiDAR as sensors. The LiDAR points are connected and transformed to the camera frame and then the object color is detected in the same area corresponding to the LiDAR points. Object detection in the camera frame is done using probabilistic color analysis. They use color discrimination approach using HSV color mapping and bayes theorem for updating the object color state. This method assumes each type of obstacle have a \textbf{unique color} (for example buoys are red) which is not the case in real scenarios. Their method was not appropriately tested and focused on small simple cases. 

\begin{figure}%
    \centering
    {\includegraphics[width=9cm, height=6cm]{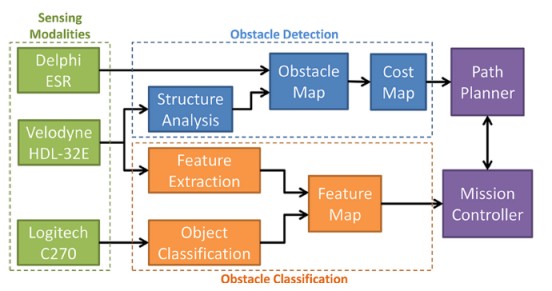} }
    \caption{Diagram of the framework used in~\cite{stanislas2018multimodal}}%
    \label{fig:fig_24}%
\end{figure}

\begin{figure}%
    \centering
    {\includegraphics[width=8cm, height=5cm]{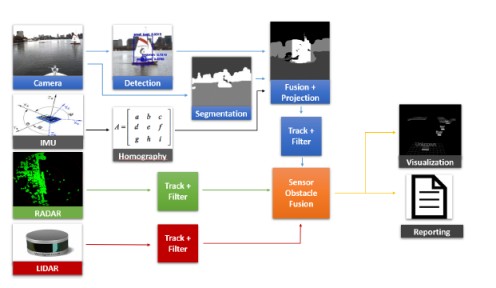} }
    \caption{System architecture used in~\cite{clunie2021development}}%
    \label{fig:fig15_1}%
\end{figure}

In~\cite{stanislas2018multimodal}, they used 3 sensors: 3D Lidar (Velodyne HDL-32E), camera (Logitech C270), and radar (Delphi ESR). LiDAR and radar are used for obstacle detection and building the obstacle map. While LiDAR and camera images are used for obstacle classification. A diagram of their framework is in figure~\ref{fig:fig_24} 

For obstacle detection using LiDAR points, they compute two defined values called \textit{stepHeight} and \textit{inclineAngle} for evey point and compare it with a predefined threshold to determine whether this point corresponds to an obstacle or not. They combine sensor cloud points coming from LiDAR and radar using probability product rule. The final costmap is computed by selecting only grid pixels with obstacle probability above a certain threshold. Camera images go through processing and blob detection operation for detecting specific classes (buoys) based on blob color and aspect ratio. The drawbacks of their method include the necessity to tune multiple parameters based on different obstacle classes and they assume unique colors for obstacles which is not the case in real scenarios. They focused only on the obstacles present in the RobotX challenge.

In~\cite{clunie2021development}, three sensors were used: marine RADAR, LIDAR, and 3 cameras, see figure~\ref{fig:fig15_1}. For the cameras, they used YOLOv4 (trained on MS-COCO dataset) for obstacles detection. For RADAR and LIDAR, PCL library is used for obstacles detection from point clusters (based on euclidean distance threshold ~\textit{D}). Land mass is filtered out by removing any clusters with an area greater than a threshold area ~\textit{A}. The optimal A and D are found empirically which is a drawback of the system. 
Land mass filtering was not completely successful and resulted in a lot of FP especially near the shore.
They developed a transformation matrix between camera images and RADAR 2D point cloud. This matrix was tuned after calibrating the camera's position with the radar's position assuming the camera and radar's locations will not be greatly shifted during the journey (not the real-world case).
For tracking objects in the camera images, they use Hungarian algorithm. For prediction, they use Kalman filter. Finally, they combine all obstacles data from all the sensors. Results of RADAR were excellent and LIDAR were good except for false detections of land obstacles. However, their tracking camera system performed badly. The calibration between camera and radar led to multiple errors.

In~\cite{wu20193d}, they preformed 3D object detection at sea using fusion between features extracted from Bird’s Eye view (BEV) point cloud coming from RADAR and images coming from camera. The two input features pass through deep neural network to output the 3D bounding boxes detections. No hardware system was developed. They tested their model on both ground and marine private datasets. However, their model is complex to be reproduced.

In~\cite{yu2021object}, they focused on the problem of moving target object tracking using photo-electric system. The drift in the position of the tracked object over the successive frames is used to control the camera to adjust its field of view towards the tracked object. The tracking is based on detection of the object in a region of interest (ROI) determined based on the object positions in the previous frames. If the tracked object cannot be detected in the ROI, the ROI area is increased, and the object is detected and matched again. This greatly improves the detection efficiency and reduces the time consumption. The prediction is carried out using Kalman Filter. The matching is implemented by correlating the detected object position in the global frame with the data coming from RADAR. As in the previous method, their model is complex to be reproduced.

\subsection{Other Approaches}

Many other approaches for obstacles detection at sea are found in the literature~\cite{park2015autonomous,woo2016vision,stateczny2019shore,ha2021radar,cane2016saliency}. 

Corner detection methods (FAST and SIFT) were adopted in ~\cite{park2015autonomous,woo2016vision} that generate features for the objects inside the image frames located under the horizon line. After that, a clustering algorithm is used to surround every group of features with bounding boxes.

In ~\cite{ha2021radar}, They used marine radar and electronic navigational chart (ENC) for generating two obstacles grid maps. The two grid maps are combined and obstacles appearing in the final grid map are classified into static or dynamic obstacles based on their appearance in successive frames.

A different approach for obstacle detection was presented in ~\cite{cane2016saliency} where they detect the obstacles at sea using saliency based algorithm. First, a saliency map is built for each frame and they perform adaptive hysteresis thresholding to locate the salient regions corresponding to potential objects. Then, they filter these candidate objects by comparing the bounding box size with a predefined threshold size formula. For tracking, they use the Hungarian algorithm and finally smooth the tracks with Kalman filter. They validated their method on the PETS 2016 maritime dataset. However, saliency based methods assume that the objects are distinguishable from their backgrounds which is not always the case.

In ~\cite{stateczny2019shore}, an algorithm for shore detection and construction using radar 3D point cloud was presented. However, an orthographic map of the environment has to be established first using a drone. 

\section{Datasets}
\begin{figure}%
    \centering
    {\includegraphics[width=8cm, height=5cm]{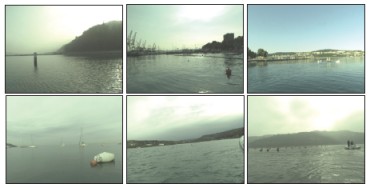} }
    \caption{MODS sample~\cite{bovcon2021mods}}%
    \label{fig:mods}%
\end{figure}

\begin{figure}%
    \centering
    {\includegraphics[width=8cm, height=7cm]{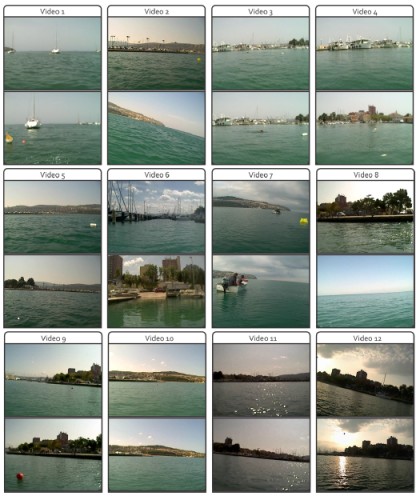} }
    \caption{MODD1 sample~\cite{kristan2015fast}}%
    \label{fig:modd1}%
\end{figure}

\begin{figure}%
    \centering
    {\includegraphics[width=8cm, height=10cm]{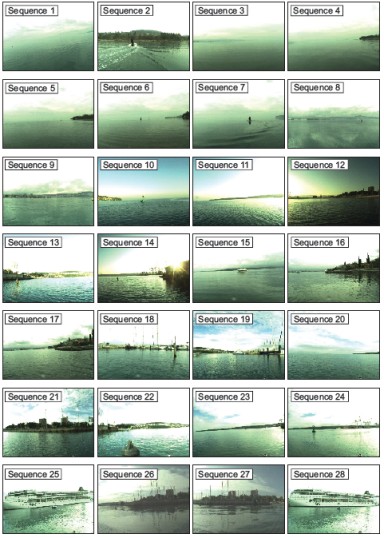} }
    \caption{MODD2~\cite{bovcon2018stereo}}%
    \label{fig:modd2}%
\end{figure}

\begin{figure}%
    \centering
    {\includegraphics[width=12cm, height=5cm]{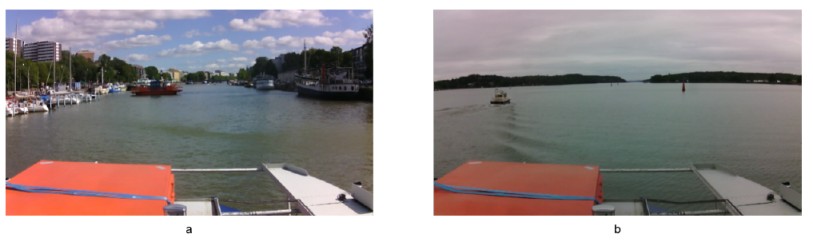} }
    \caption{ABOships sample~\cite{iancu2021aboships}}%
    \label{fig:abo}%
\end{figure}

\begin{figure}%
    \centering
    {\includegraphics[width=10cm, height=5cm]{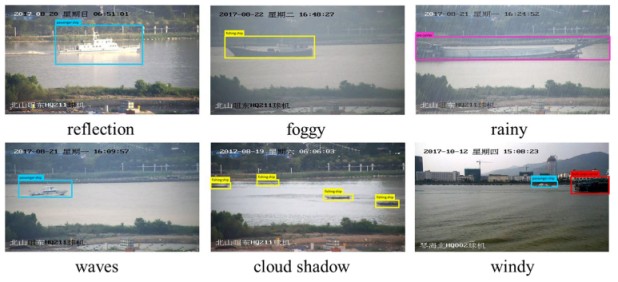} }
    \caption{SeaShips sample~\cite{shao2018seaships}}%
    \label{fig:seaships}%
\end{figure}

\begin{figure}%
    \centering
    \subfloat{{\includegraphics[width=6cm, height=4cm]{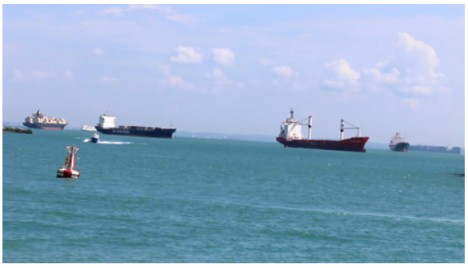} }}%
    \qquad
    \subfloat{{\includegraphics[width=5cm, height=4cm]{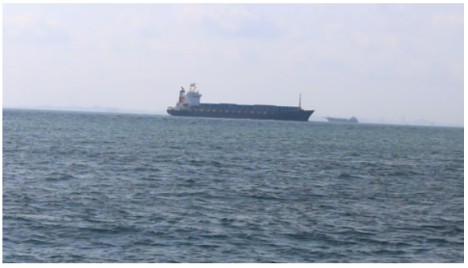} }}%
    \caption{SMD sample~\cite{SMD}}%
    \label{fig:smd}%
\end{figure}

\begin{figure}%
    \centering
    \subfloat{{\includegraphics[width=6cm, height=4cm]{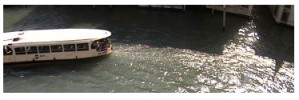} }}%
    \qquad
    \subfloat{{\includegraphics[width=6cm, height=4cm]{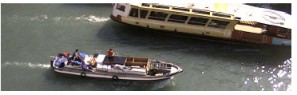} }}%
    \caption{MarDCT sample~\cite{bloisi2015argos}}%
    \label{fig:mardct}%
\end{figure}

\begin{figure}%
    \centering
    {\includegraphics[width=8cm, height=7cm]{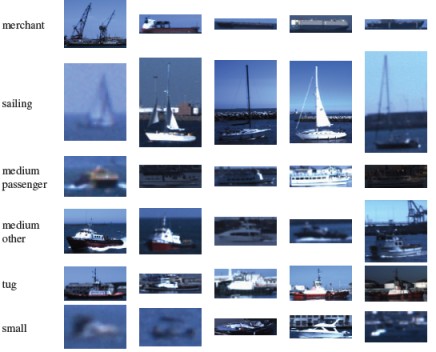} }
    \caption{VAIS sample~\cite{zhang2015vais}}%
    \label{fig:vais}%
\end{figure}

\begin{figure}%
    \centering
    {\includegraphics[width=8cm, height=5cm]{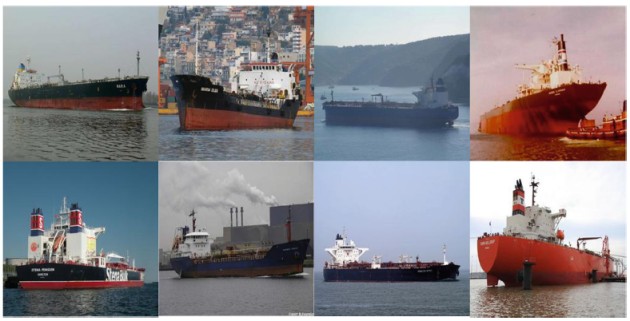} }
    \caption{MARVEL sample~\cite{gundogdu2016marvel}}%
    \label{fig:marvel}%
\end{figure}

\begin{figure}%
    \centering
    {\includegraphics[width=8cm, height=9cm]{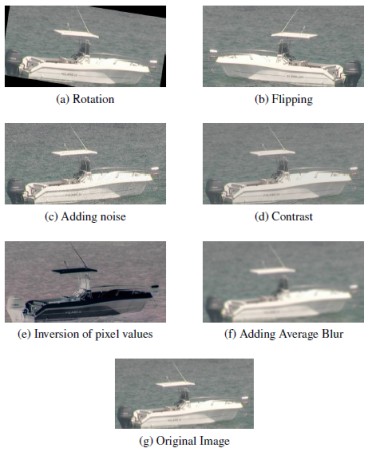} }
    \caption{Boat Re-ID sample~\cite{spagnolo2019new}}%
    \label{fig:boatreid}%
\end{figure}

\begin{figure}%
    \centering
    {\includegraphics[width=14cm, height=3cm]{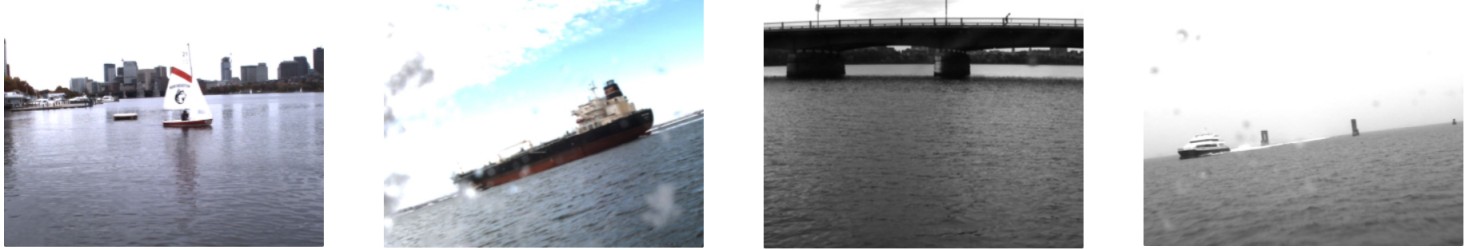} }
    \caption{MIT dataset sample~\cite{MIT}}%
    \label{fig:mit}%
\end{figure}

\subsection{Annotated Datasets}
There are multiple publicly available ~\textbf{annotated} image datasets for object detection in marine environments. Some of them are sequence based (sequences of consecutive frames taken from video) and some of them are image based (uncorrelated frames). Sequence based datasets are especially useful for testing and evaluating the tracking module. However, image-based datasets are only useful for training and testing the detection module.

Table\ref{table_datasets} presents a comparison between the investigated datasets that will be presented in the following subsections. BB is an abbreviation for bounding boxes.

There are also other recently released marine datasets~\cite{MID,shao2021glsd,zheng2020mcships,patino2016pets}. However, they are currently private.

\subsubsection{Sequence-Based Datasets}
\begin{enumerate}
    
    \item \textbf{MODS}\cite{bovcon2021mods}: A dataset of 81k images. Their dataset annotations were performed in a very strict procedure to ensure quality and accuracy. Their obstacles classes are 3: Vessel, Person, Other. Most of the obstacles present in the dataset are in the range of 15-20 meters away from the USV. It was acquired on the Slovenian coast.
    
    \textbf{Dataset Sample}: See figure ~\ref{fig:mods}
    
    
    \item \textbf{MODD1}\cite{kristan2015fast}: A dataset of twelve fully annotated videos with semantic segmentation masks as well as obstacle bounding boxes. It was acquired in the gulf of Trieste in the port of Koper, Slovenia. 
    
    \textbf{Dataset Sample}: See figure ~\ref{fig:modd1}
    
    
    \item \textbf{MODD2}\cite{bovcon2018stereo}: A large dataset recorded over several months with various weather conditions. The video frames are annotated with semantic segmentation masks as well as obstacle bounding boxes. The dataset was acquired in the gulf of Trieste in the port of Koper, Slovenia.
    
    \textbf{Dataset Sample}: See figure ~\ref{fig:modd2}
    
    
    \item \textbf{ABOships}\cite{iancu2021aboships}: A marine dataset of around 10k images. The annotation of objects belongs to one of 11 categories: boat, cargoship, cruiseship, ferry, militaryship, miscboat, miscellaneous (floater), motorboat, passengership, sailboat and seamark. The frame rate is very low (1 frame each 15 sec). The top surface of the watercraft appears in all the frames of the dataset. It was acquired in Aura river and port of Turku in Finland.

    \textbf{Dataset Sample}: See figure ~\ref{fig:abo}
    
    
    \item \textbf{Singapore Maritime Dataset (SMD)}\cite{SMD}: dataset of optical and IR video frames at various times of the day and weather conditions. The obstacles are annotated with bounding boxes. Each obstacle is further labelled with three types: 
    \begin{itemize}
        \item Distance type: Near, Far, Other
        \item Object Type: Ferry, Buoy, Vessel, Speed Boat, Boat, Kayak, Sail Boat, Swimming Person, Flying Bird, Other
        \item Motion Type: Moving, Stationary, Other
    \end{itemize}
    
    \textbf{Dataset Sample}: See figure ~\ref{fig:smd}
    
\end{enumerate} 

\subsubsection{Image-Based Datasets}
\begin{enumerate}
    \item \textbf{SeaShips}\cite{shao2018seaships}: It contains annotated images of sea ships of 6 classes (ore carrier, bulk cargo carrier, general cargo ship, container ship, fishing boat, and passenger ship) with their bounding boxes in the Pascal VOC format. The images are taken from the shore around the Hengqin Island, Zhuhai city, China in 2018 and they cover different backgrounds. 
    
    \textbf{Dataset Sample}: See figure ~\ref{fig:seaships}
    

    
    
    \item \textbf{MarDCT}\cite{bloisi2015argos}: Top view images of different boats, with each image containing only one boat in the center and it has a bounding box and a classification label. They also provide identification number for each boat for tracking purposes. The dataset was acquired by their ARGOS system operating 24/7 in Venice, Italy. It has 24 specific categories of the boats and these can be further grouped to 5 generalized categories.
    
    \textbf{Dataset Sample}: See figure ~\ref{fig:mardct}
    
    
    \item \textbf{VAIS}\cite{zhang2015vais}: A small dataset (1500 images) of ships images annotated with their bounding boxes and class. There are 6 ship classes: merchant, sailing, passenger, medium, tug, and small. There are paired visible and IR images in this dataset. They found that classification of images at night is more accurate on IR images and established a benchmark for that. The dataset was captured over nine days and annotated them manually. They officially divided their dataset into training and testing splits to unify the evaluation on different object classification models.
   
   \textbf{Dataset Sample}: See figure ~\ref{fig:vais}
   
    
    \item \textbf{MARVEL}\cite{gundogdu2016marvel}: A large size dataset of marine vessels images. The images are collected from the internet. Each image contain only one vessel and there is no bounding box annotations available. Each image of vessel has one class of 26 possible ship classes. This can be used for image classification purposes but it is not quite useful for our detection and tracking system. 
    
    \textbf{Dataset Sample}: See figure ~\ref{fig:marvel}
    
    
    \item \textbf{Boat Re-ID}\cite{spagnolo2019new}: It is composed by annotated images of boats. The images are cropped and cover different types of boats with different view angles, orientations, and noises. It also contains the ID of the boats, because each boat can have multiple images with different noise or view angle. It can be used for boat re-identification (or, in our case, for testing the appearance matching function of the tracker). The datasets were acquired in Porto Cesareo, Italy in 2019.
    
    \textbf{Dataset Sample}: See figure ~\ref{fig:boatreid}
    
\end{enumerate}

\begin{table}[h]
\begin{center}
\caption{Available Maritime Image Datasets of Ships}
\label{table_datasets}
\resizebox{\linewidth}{!}{
\begin{tabular}{c c c c c c c}
\hline
\textbf{Dataset} & \textbf{Shooting} & \textbf{Resolution} & \textbf{Camera Type} & \textbf{Size} & \textbf{Classes} & \textbf{Annotation}\\
\hline

MODS\cite{bovcon2021mods} & on-board & 1278$\times$958 & Vrmagic & 81k & 3 & BB\\
MODD1\cite{kristan2015fast} & on-board & 640$\times$480 & Axis 207W & 4.5k & - & BB\\
MODD2\cite{bovcon2018stereo} & on-board & 1278$\times$958 & Vrmagic & 11.5k & - & BB\\
ABOships\cite{iancu2021aboships} & on-board & 1280$\times$720 & HDcamera & 10k & 11 & BB\\
SeaShips\cite{shao2018seaships}   & on-shore & 1920$\times$1080 & variant & 31k & 6 & BB\\
SMD\cite{SMD}   & both & 1080$\times$1920 & Canon 70D & 17k & 10 & BB, tracks\\
MarDCT\cite{bloisi2015argos}   & on-shore top & 800$\times$240 & variant & 12k & 5 & BB, id\\
VAIS\cite{zhang2015vais}   & on-shore & 5056$\times$5056 & ISVIIC-C25 & 1.5k & 6 & BB, id\\
MARVEL\cite{gundogdu2016marvel}   & on-shore & variant & variant & 140k & 26 & id\\
Boat Re-ID\cite{spagnolo2019new}   & on-shore & variant & AXIS Q6035-E & 5.5k & - & id\\
\hline

\end{tabular}
}
\end{center}
\end{table}

\subsection{Non-Annotated Datasets}
Autonomous Underwater Vehicles Laboratory (AUV Lab) at MIT has recently published a non-annotated multi-modal sensor dataset for mobile robotics research in the marine domain~\cite{MIT}. The dataset is composed of videos acquired from optical cameras and IR cameras, point clouds acquired from 3D LiDAR, RADAR images, and IMU data. All these data are synchronized together and recorded in the form of ROS bag files. The data was acquired during different missions in the sea with variant number of obstacles and weather conditions. The data was collected on the Charles River, Boston, MA out into Boston Harbor.

The sensors used in their system are:
\begin{itemize}
    \item VLP-16 Velodyne 3D Lidar
    \item 4G Broadband Radar
    \item 3 FLIR Blackfly 1.3 MP Cameras
    \item 2 FLIR ADK 0.3 MP Infrared Cameras
\end{itemize}

Although this dataset is not annotated, it provides a strong base for testing marine object detection systems and evaluate their performance. By manually annotating a small subset of the data, we can test the behaviour of our developed object detection and tracking system on realistic and challenging data. It is also an ideal test bed for our overall (camera+3D LiDAR) pipeline. 

\textbf{Dataset Sample}: See figure ~\ref{fig:mit}


\section{Robotics Operating System (ROS)}
\begin{figure}%
    \centering
    {\includegraphics[width=7cm, height=5cm]{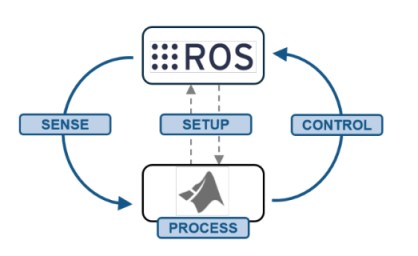} }
    \caption{ROS ecosystem}%
    \label{fig:ros}%
\end{figure}

ROS~\cite{quigley2009ros} is a flexible platform to build robotics software applications. Its collection of tools, libraries and conventions greatly simplifies the task of building complex and robust robotics behaviours. In addition, ROS was created to encourage collaborative robotics software development across the world. The ecosystem of ROS is illustrated in figure ~\ref{fig:ros}.

The file system and nodes representation in ROS are extremely helpful in organizing and building robotics tasks. ROS offers a message passing interface that provides inter-process communication referred to as middleware. The middleware provides facilities like: publish/subscribe anonymous message passing, recording and playback of messages, remote procedure calls, and distributed parameter system. In addition, ROS provides common robot-specific features that help in running basic and core robotics functions. The offered features include standard
message definitions for robots and sensors, robot geometry library and pose estimation tools. 

Perhaps the most well-known tool in ROS is Rviz. Rviz provides general purpose, three-dimensional visualization of many sensor data types and any URDF-described robot. We can easily visualize the laser scanned data, robot’s odometry, received video frames, and many other topics that the robot subscribes to. Rviz can be seen as a tool to visualize what your robot can see. Another useful tool in ROS is rqt. Using the rqt graph plugin we can introspect and visualize a live ROS system, showing nodes and the connections between them, and being able to easily debug and understand our running system and how it is structured.

For all the mentioned features, we have chosen ROS as a software platform to develop and test our robot’s navigation system on. The ROS version we use is ROS noetic which is the 13th official ROS release and is supported on our Ubuntu 20.04 system.

\chapter{Methodology}
\label{chap:second}
\ifpdf
    \graphicspath{{Chapter2/Figures/PNG/}{Chapter2/Figures/PDF/}{Chapter2/Figures/}}
\else
    \graphicspath{{Chapter2/Figures/EPS/}{Chapter2/Figures/}}
\fi

In this chapter, the multi-modal obstacle avoidance methodology that was developed previously will be explained in details.

\begin{figure}%
    \centering
    {\includegraphics[width=10cm, height=4cm]{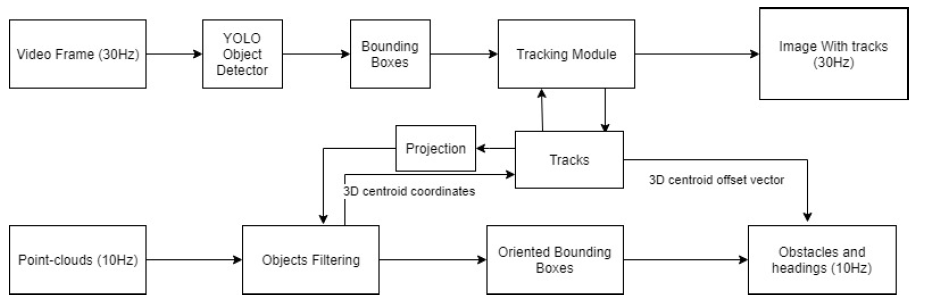} }
    \caption{Block Diagram of the Pipeline}%
    \label{fig:block}%
\end{figure}

Figure ~\ref{fig:block} depicts a block diagram of the pipeline elements. The input to the system is the video frames and the lidar point clouds perceived one at a time. The output is the locations and headings of the detected obstacles in the 3D reference frame. 

\section{Obstacle Detection}
Images offer rich semantics compared to range measurements and information about the type of obstacles can be extracted from images. We run a per-frame object detector YOLOv3 on the incoming video frames to extract the obstacles classes (for now it is limited to boats) and their bounding boxes. Relying on video-oriented object detector was not possible because the available video datasets that can be used for training such object detectors do not contain the 'boat' class. Therefore, the image-based object detector is used and optimized to be used for a video scenario.

Below a detailed description of the selected detector is provided highlighting the main advantages to the base architecture YOLO~\cite{redmon2016you}. 

\subsection{YOLOv3 Architecture}

Unlike YOLO, YOLOv3 uses a fully convolutional neural network following the proposal free object detection architecture. Inspired by ResNet-101 residual connections, YOLOv3 proposes a customized network called the feature extractor (referred to as DarkNet). The feature extractor is pre-trained on ImageNet classification task (a typical option to train network backbones). 

YOLOv3 features several enhancements that augment the accuracy of the detector (some are present in YOLO2 too):
\begin{itemize}
    \item  \textbf{Aspect ratios:} the network assumes a set of frequent aspect ratios (called anchor boxes) for the objects. This facilitates the task of bounding boxes prediction to predicting an offset of the anchor box. \item Multi-Class prediction: YOLOv3 replaces the final softmax activation function present in YOLO with separate logistic activations for each of the classes. This allows for multi-class prediction (e.g. a single anchor box can predict the presence of both a pedestrian and a child). Each of the bounding boxes detected within a frame has the full set of class probabilities (80 in the case of MS COCO~\cite{lin2014microsoft}).
    \item \textbf{Offset constraining:} A sigmoid activation function is used to constrain the values of the bounding boxes offset to avoid randomizing them. It is also good for stabilizing early training iterations.
    \item \textbf{Multi-scale predictions:} the predictions are extracted using three different layers each with a unique spatial resolution to detect objects with various sizes.
    
\end{itemize}

Output predictions from the network that have class score above a threshold (the threshold used by the authors is 0.5) are fed to a non-maximum suppression (NMS) module. The module discards all predictions that have IoU (intersection over union) overlap of 0.45 or higher with other output predictions of the same class. 

The real-time performance of YOLOv3 is relatively high compared to other detectors and the network is able to give predictions with close accuracy to other slower models (based on region proposal).

The used YOLOv3 here is previously trained on detection task of MS COCO dataset. Fine-tuning such network requires training marine image dataset of big size and cover variable conditions. Such datasets were listed in the previous chapter and we intend to use them for fine-tuning in future work.

We decided to set the detection threshold to 0 , unlike the typical choice of 0.5 (the threshold is used to filter out predictions that have : ( \(class\_prob * objectness\_score < threshold\) ). The intuition behind this is that the tracking module follows the detection module, and, in any case, it can benefit from correct detections with low confidence while wrong detections can be handled by the tracking procedure that will be explained later.

\section{Obstacle Tracking}

\begin{figure}%
    \centering
    {\includegraphics[width=14cm, height=6cm]{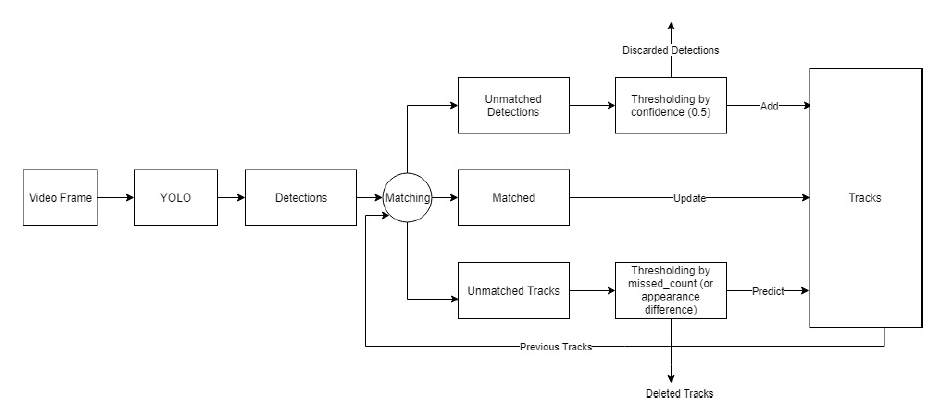} }
    \caption{Proposed tracking pipeline}%
    \label{fig:track_pip}%
\end{figure}

In figure~\ref{fig:track_pip}, the overall tracking procedure is summarized. 

\begin{figure}%
    \centering
    {\includegraphics[width=14cm, height=6cm]{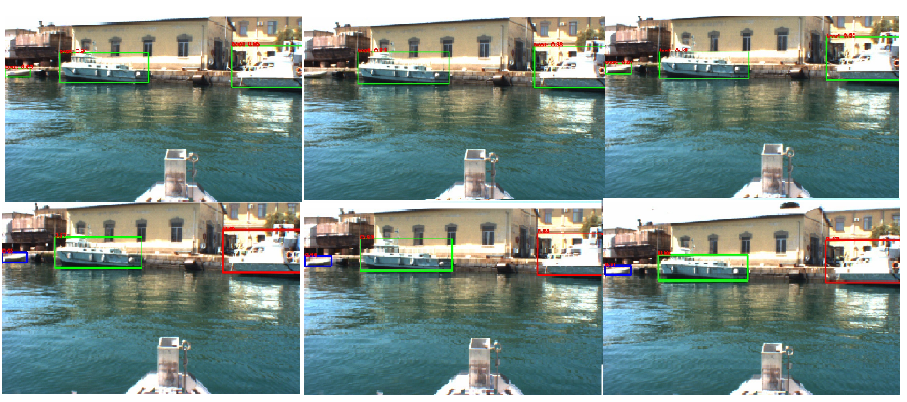} }
    \caption{Difference between detection and tracking: showing frames with results of YOLO detections (top), and the proposed tracking results (bottom).}%
    \label{fig:det_track}%
\end{figure}

In our work, the tracking is guided by both outputs of the object detection network and predictions from state prediction algorithm. Using such method, the tracker is able to account for false-negative and false-positive predictions incurred by the object detector. The tracker recognizes objects across video frames and assign each of them a tracking ID. In figure~\ref{fig:det_track}, a comparison between the outputs of the detector and the tracker is depicted. 

\subsection{Data Association}

One of the core problems in tracking is the data association problem. Data association means defining the attributes to decide how tracked objects are matched with detected ones. In our work we rely on the Hungarian algorithm~\cite{kuhn1955hungarian} to perform optimal minimum cost matching between tracks and detections. We define the matching cost to depend on both location and appearance attributes and to have three components as follows: 

$$Matching Cost = ConfidencePenalty * (IoU\_dist + desc\_dist)$$
 
where:
\begin{itemize}
    \item $confidencePenalty$ is the penalty of low confidence detections (below 0.5) and is equal to the prediction uncertainty (1-confidence). This is to differentiate between the matching cost of low-confidence detections and high-confidence detections. 
    \item $IoU\_dist$ is the distance in the image plane between the two bounding boxes belonging to the track and the detected object and it equals 1-IoU (known as Jaccard distance). 
    $$d_{j}(A,B) = \dfrac{|A \cup B| - |A \cap B|}{|A \cup B|}$$
    \item $desc\_dist$ is the distance measured in an appearance representation space. The representation can be any size-agnostic hand-crafted or learned one. We found the Histogram of Oriented Gradients (HOG) with number of blocks (2 by 2) that yields a representational histogram of 36 bins. We measure the distance between two representations using Manhattan distance (L1).
\end{itemize}

\subsection{Tracks Addition and Deletion Policy}

There is a trade-off in deciding the time window for tracking unmatched tracks. If we predict unmatched tracks for a long time window, this will cause eventually errors in the prediction and high false positive rate. On the other hand, if we use short time window, this may make the unmatched tracked objects not persist for long time and this can increase the false negative rate. 

To construct a good policy, we shall take into consideration several factors:
\begin{itemize}
    \item \textbf{The detector:} if the detection neural network has high false negative rate (many objects are missed in the detection output), this will require a relaxed track deletion policy. If the detector has high false positive rate, this will require restricted track addition policy.
    \item \textbf{The prediction model:} the accuracy of the prediction model has to be taken into consideration in deciding the time window for predicting unmatched tracks.
    \item \textbf{The environment:} in crowded scenes, predicting tracks for relatively long times can make them overlap with other detections of other objects and this can lead to identity switch. Also, in some environmental conditions like rain and winds the video camera gets shaked continuously and this complicate the prediction procedure.
    
\end{itemize}

After analyzing all the previous factors and the available detector and predictor options, we adopt the following policy:  
\begin{itemize}
    \item \textbf{Track Addition:} detections are added to the tracks list only if they have high confidence (above 0.5), and a 3-frames period is required of persistent detections before the track is outputted by the tracker. An automatic Id is assigned of the track when it is added (using a counter variable).
    \item \textbf{Track deletion:} tracks get deleted if they are missed for 10 consecutive frames or if the distance between the representation of the track in the current frame and the last seen matched detection, is larger than some threshold.
\end{itemize}

\subsection{Prediction Model}
The main rule of the prediction model is to predict the new position of a track in the current frame in case it is not matched with any of the detections. This prediction is calculated based on the previous history of the track positions. 

The typical yet effective prediction model that we choose to use in our system is Kalman Filter. Kalman filters use a combination of predefined motion model and a sequence of measurement updates to estimate the next most probable location of the object. 

Below, we provide more details on how we utilize Kalman Filter to derive
predictions.

\subsubsection{Kalman Filter}

Simple Kalman filters and Extended Kalman filters (EKF) are used to handle linear and non-linear state transitions, respectively. We focus in this work on simple Kalman filters due to the favored mathematical and computational tractability.

We adopt a constant acceleration motion model where the model predicts a variable velocity every time step. The measurements are the locations of two corners of the bounding box (top-left and bottom-right corners).

Formally, the system state is represented with the vector $x_k$ that is formed of the coordinates of the two corners and corresponding velocities and accelerations:

$$x_k = [x_{min}, y_{min}, x_{max}, y_{max}, v_{xmin}, v_{ymin}, v_{xmax}, v_{ymax}, a_{xmin}, a_{ymin}, a_{xmax}, a_{ymax}]^T$$

While the measurement vector is defined as:

$$z_k = [x_{min}, y_{min}, x_{max}, y_{max}]^T$$

The transition matrix $A$ satisfies the constant velocity model, and is given by ( where $\Delta t$ is assumed to equal 1):

$$A = \begin{bmatrix}
1 & 0 & 0 & 0 & \Delta t & 0 & 0 & 0 & \dfrac{{\Delta t}^2}{2} & 0 & 0 & 0\\
0 & 1 & 0 & 0 & 0 & \Delta t & 0 & 0 & 0 & \dfrac{{\Delta t}^2}{2} & 0 & 0\\
0 & 0 & 1 & 0 & 0 & 0 & \Delta t & 0 & 0 & 0 & \dfrac{{\Delta t}^2}{2} & 0\\
0 & 0 & 0 & 1 & 0 & 0 & 0 & \Delta t & 0 & 0 & 0 & \dfrac{{\Delta t}^2}{2}\\
0 & 0 & 0 & 0 & 1 & 0 & 0 & 0 & 0 & 0 & 0 & 0\\
0 & 0 & 0 & 0 & 0 & 1 & 0 & 0 & 0 & 0 & 0 & 0\\
0 & 0 & 0 & 0 & 0 & 0 & 1 & 0 & 0 & 0 & 0 & 0\\
0 & 0 & 0 & 0 & 0 & 0 & 0 & 1 & 0 & 0 & 0 & 0
\end{bmatrix}$$

While the observation matrix $H$ is defined as:

$$H = \begin{bmatrix}
1 & 0 & 0 & 0 & 0 & 0 & 0 & 0 & 0 & 0 & 0 & 0\\
0 & 1 & 0 & 0 & 0 & 0 & 0 & 0 & 0 & 0 & 0 & 0\\
0 & 0 & 1 & 0 & 0 & 0 & 0 & 0 & 0 & 0 & 0 & 0\\
0 & 0 & 0 & 1 & 0 & 0 & 0 & 0 & 0 & 0 & 0 & 0
\end{bmatrix}$$

A filter instance is created for each tracked object and the predictions are used to predict the box position of the tracked object in case of missing detection. The Kalman filter output is used only after 10 frames of tracking to avoid the errors present in the initial iterations.

\section{Sensor Fusion}

In order to acquire 3D information about the surrounding environment, we rely on LiDAR range sensor that can cover around 100 meters of 360 degrees horizontally and 30 degree vertically under different weather conditions. Historically, RADAR was the natural choice as a range sensor for its long range capabilities that are especially useful in offshore marine applications. However, LiDAR is preferred to RADAR in near-port marine applications due to its richer 3D information compared to RADAR.

The inputs coming from the two sensor modalities (video camera and LiDAR) are synchronized and related by a transformation matrix. The transformation matrix project the points in the 3D reference frame received from the LiDAR to pixels in the camera image plane. Such transformation matrix can be computed using a calibration procedure that aims at finding a projection matrix that transforms points received from the LiDAR in the 3D reference frame to points in the image plane. The equation of such projection can be expressed as:

$$P_{img}(x, y) = Proj\_Matrix * P_{lidar}(X,Y,Z)$$

Such projection matrix is already available for the USV that collected the data in the MIT dataset. However, eventually, when we build our own hardware system, we will need to compute such matrix. 

\subsection{3D Localization of Obstacles}

To localize the detected objects in the 3D reference frame, the intuitive approach would be to project the point cloud on the image plane and filter points inside the bounding boxes of the detected objects and these will be the 3D points corresponding to the detected objects. However, this approach is computationally expensive due to the huge number of points received from the LiDAR. An alternative approach, that we follow, is to localize the bounding boxes in the 3D reference frame using the pseudo-inverse of the projection matrix. The projection of the bounding box in the 3D frame would correspond to a pyramid shape. We use the generated pyramids to filter points in the point cloud and extract the clusters of points corresponding to the detected objects. 

Mathematically speaking, the transformation matrix between the 3D reference frame and the image plane is a non-square matrix (since the number of dimensions of the source and destination spaces do not match) and can not be inverted. However, by using the pseudo-inverse (obtained using the Moore-Penrose inverse~\cite{moore1920reciprocal}), we can obtain a transformation matrix that projects 2D points to a set of 3D points aligned on a 3D line. 

The geometrical intuition behind the non-invertible projection matrix is that projecting a 3D point onto a plane (image-plane in our situation) discards information about depth (distance) to the plane. In fact, an infinite number of points with different depths will project to the same 2D point on the image plane. Thus, using the pseudo-inverse will, similarly, produce an infinite number of 3D points aligned on a line. Consequently, projecting the four corners of a bounding box on the 3D reference frame yields four lines that intersect in the origin and diverge as they move away from the origin. 

Each consecutive two lines define a plane (total of four planes). We can estimate the equations of the planes using the cross product of the two lines. After that, 3D points filtering is done using simple geometry; the 3D point coordinates are plugged into the equations of the four planes and agreeing signs indicate that the point is completely contained with the volume bounded by the four planes.

Next, a clustering algorithm is used to cluster the filtered points from the pyramid and we extract the biggest cluster to be the one corresponding to the detected object. Another approach would be to consider the obstacle to be the nearest cluster of points.

After that, the extracted clusters are projected on the 2D sea-surface plane (since we are not interested in the height of the obstacles). The projection is achieved by making use of the measurements received from the IMU sensor mounted on the USV that contain the orientation (in the form of quaternion) of the USV with respect to the reference frame of the sea-surface.

\begin{figure}%
    \centering
    {\includegraphics[width=14cm, height=7.6cm]{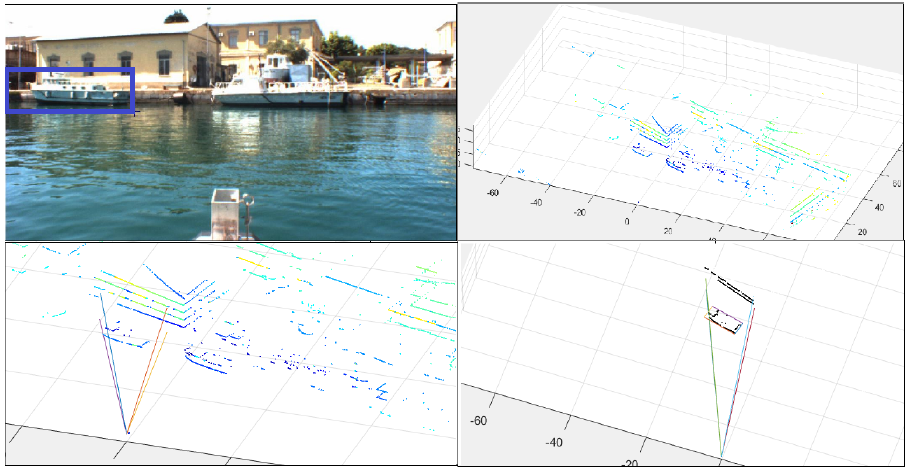} }
    \caption{3D Localization steps: a track bounding box (top-left), the corresponding point cloud input (top-right), the projected bounding box on the 3D reference frame (bottom left), Filtering using the 2D bounding box of the obstacle.}%
    \label{fig:loc}%
\end{figure}

After detecting the cluster of points belonging to the obstacle, a simple procedure is followed to construct a 2D oriented bounding box in the 3D reference frame. This is done by applying PCA to the points (after projecting them to the sea-surface plane) to estimate the two dimensions of the largest variance (see Figure~\ref{fig:loc} for an example of possible input and the corresponding output).

\chapter{Software Architecture}
\label{chap:third}
\ifpdf
    \graphicspath{{Chapter3/Figures/PNG/}{Chapter3/Figures/PDF/}{Chapter3/Figures/}}
\else
    \graphicspath{{Chapter3/Figures/EPS/}{Chapter3/Figures/}}
\fi

In this chapter, the software architecture on ROS of the methodology described in the previous chapter will be demonstrated. Nevertheless, the methodology was adjusted during the software implementation after the analysis of the results obtained from testing the system on the MIT dataset. These results will be presented and analyzed in the following chapter. 

The major modifications on the methodology described previously are the following:

\begin{itemize}
    \item \textbf{The tracking of obstacles on the image plane was omitted} and replaced with a simpler tracking procedure on the 3D plane of the point cloud clusters of obstacles. This tracking procedure relies solely on matching obstacles across time frames by comparing the distance between the centers of their enclosing 3D bounding boxes.
    \item In addition to the detection and tracking of obstacles that lie within the camera view, \textbf{another module was added to detect and track surrounding obstacles outside the camera view (with lower confidence).} The procedure detects obstacles outside the camera view using only the Lidar point cloud. For the detection, it relies solely on the clustering of the received point cloud to extract clusters of points corresponding to obstacles and enclose them with 3D bounding boxes. For the tracking, it uses the same method for tracking obstacles within the camera view that is described above.
\end{itemize}

The software was developed and tested on ROS noetic that is installed on a PC with ubuntu 20.04 operating system and Nvidia RTX2 GPU. The code is currently private and can be accessed after permission \footnote{\url{ https://bitbucket.org/isme_robotics/obstacle_detection_tracking/src/master/}}

In the coming sections, the UML diagrams representing the software architecture will be presented and explained as well as the system features and working assumptions.

\section{Component Diagram}

\begin{figure}%
    \centering
    {\includegraphics[width=14cm, height=18.5cm]{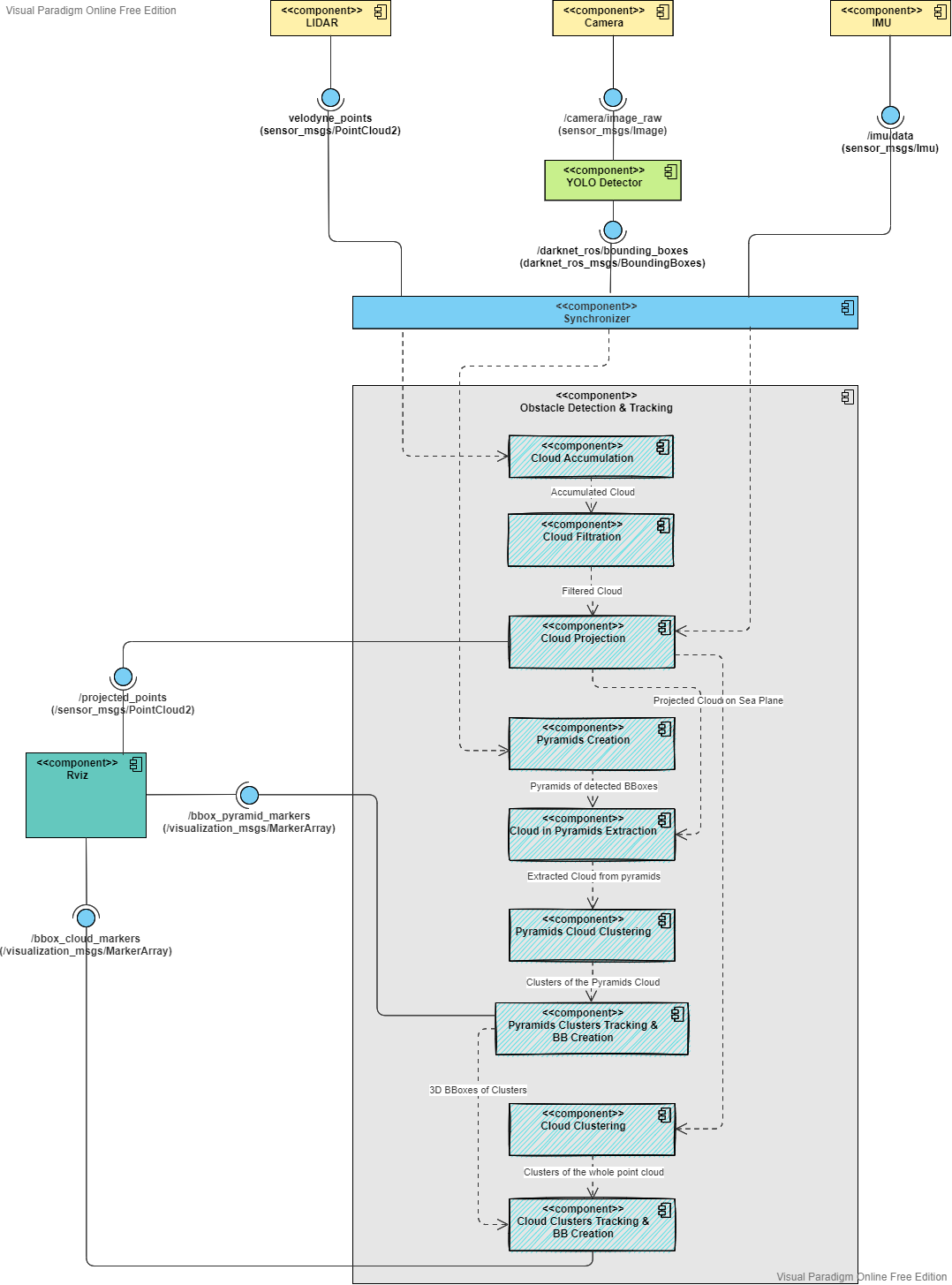} }
    \caption{System Component Diagram. BB is an abbreviation for bounding boxes}%
    \label{fig:comp_diag}%
\end{figure}

The component diagram of the developed software is depicted in figure ~\ref{fig:comp_diag}. It is worth clarifying that the term "component" here doesn't have to correspond to a hardware component or a ROS node. However, it can be used to represent a group of functions or operations that does a specific task.

The inputs to the system are the time-stamped data perceived from the three sensors we use: 
\begin{itemize}
    \item \textbf{3D LIDAR:} it publishes 3D scans of the surrounding environment in the form of point cloud. The point cloud is received continuously (with 10Hz rate in the case of MIT dataset) on the ROS topic \textit{"velodyne\_points"} as a ROS message of type \textit{"sensor\_msgs/PointCloud2"}
    \item \textbf{Camera:} it publishes video frames of the view in front of the camera. Depending on where the camera is installed on the vehicle, the view changes and the obstacles that can be detected in the images change accordingly. The images are received on the ROS topic \textit{"/camera/image\_raw"} as a ROS message of type \textit{"sensor\_msgs/Image"}
    \item \textbf{IMU:} We use the IMU to get the orientation of the vehicle with respect to the sea surface frame. This is useful in order to project the point cloud correctly on the sea surface plane as will be explained later. The IMU data is received on the ROS topic \textit{"/imu/data"} as a ROS message of type \textit{"sensor\_msgs/Imu"}
\end{itemize}

Currently, a brief description of each component in the software will be given including what it does, its inputs, and its outputs. 

\begin{itemize}
    \item \textbf{YOLO Detector:} This component implements real-time YOLOv3 object detection on ROS. The code is modified from the package darknet\_ros\footnote{\url{https://github.com/leggedrobotics/darknet_ros}} where the network is already pre-trained on COCO and PASCAL-VOC datasets. It subscribes to the camera images and, then, publishes a list of bounding boxes of the detected objects of class "Boat". The bounding boxes are published on the topic \textit{"/darknet\_ros/bounding\_boxes"} as a message of type \textit{"darknet\_ros\_msgs/BoundingBoxes"}. The published messages include the time stamp, the locations of the bounding boxes, and their dimensions. The time stamp is modified to be the same time stamp of the received camera image. This is done for synchronization purposes.
    \item \textbf{Synchronizer:} The synchronizer is mainly responsible for synchronizing the point cloud, IMU, and bounding boxes messages based on their time stamps to output them in the form of a group of 3 synchrnoized messages (point cloud, bounding boxes, imu) one at a time. The messages contained in each group have time stamps close to each other so they approximately correspond data acquired in the same time instance.
    \item \textbf{Obstacle Detection \& Tracking:} This is the main component where the algorithm is implemented. The inputs to this component are the synchronized groups of messages, and the outputs are the visualization messages to Rviz. Each group of synchronized messages pass through the operations inside this component to detect and track the obstacles in the 3D reference frame. 
    
    The sequence of operations inside this component are the following: 
    
    \begin{enumerate}
        \item \textbf{Cloud Accumulation:} Due to the sparcity of the received point cloud, we accumulate the current point cloud to previously received clouds to form a more-dense and representing cloud. The number of previous clouds to accumulate has to be selected reasonably in accordance with the point cloud reception rate. For example, in the MIT dataset, the point cloud rate is 10Hz. A reasonable number for the accumulation window can be 10 (or less). In this case, we are adding point clouds received in the last second. During one second of time, there will not be significant changes in the locations of the obstacles and, therefore, the accumulated dense cloud will be approximately representing of the current time instance. 
        \item \textbf{Cloud Filtration:} This block filters the point cloud to remove points that are reflected from the USV itself and also removes the far-away points.
        \item \textbf{Cloud Projection:} Given the orientation of the USV present in the IMU data, this block projects the point cloud on the sea surface plane. Then, it publishes this projected cloud to Rviz to be visualized.
        \item \textbf{Pyramids Creation:} It creates the pyramids corresponding to the bounding boxes detected by YOLO.  
        \item \textbf{Cloud in Pyramids Extraction:} It extracts the points inside the pyramids.
        \item \textbf{Pyramids Cloud Clustering:} It clusters the point cloud extracted from the pyramids to identify the obstacles clusters in 3D.
        \item \textbf{Pyramids Clusters Tracking \& BB Creation:} This block assigns the tracking IDs to the clusters of obstacles extracted from the pyramids using the tracking-by-distance approach described before. It creates the enclosing 3D bounding boxes of the clusters and gives them unique colors based on the tracking ID. Then, it publishes them to Rviz to be visualized.
        \item \textbf{Cloud Clustering:} This block clusters the whole point cloud to detect obstacles not lying within the camera view.
        \item \textbf{Cloud Clusters Tracking \& BB Creation:} As before, this block assigns the tracking IDs to the clusters of obstacles extracted from the whole point cloud using the tracking-by-distance approach. It creates the enclosing 3D bounding boxes of the clusters and gives them unique colors based on the tracking ID. Then, it publishes them to Rviz to be visualized. It can also use the bounding boxes extracted from the pyramids to assign labels indicating the obstacles type on the bounding boxes lying within the camera view. The obstacles not lying within the camera view will have no label (unknown).
    \end{enumerate}
    
    \item \textbf{Rviz:} a 3D visualization software for ROS that can visualize different types of ROS messages including point cloud, images, and markers (text, boxes, spheres, etc).
\end{itemize}

\section{Class Diagram}
\begin{figure}%
    \centering
    {\includegraphics[width=14cm, height=8cm]{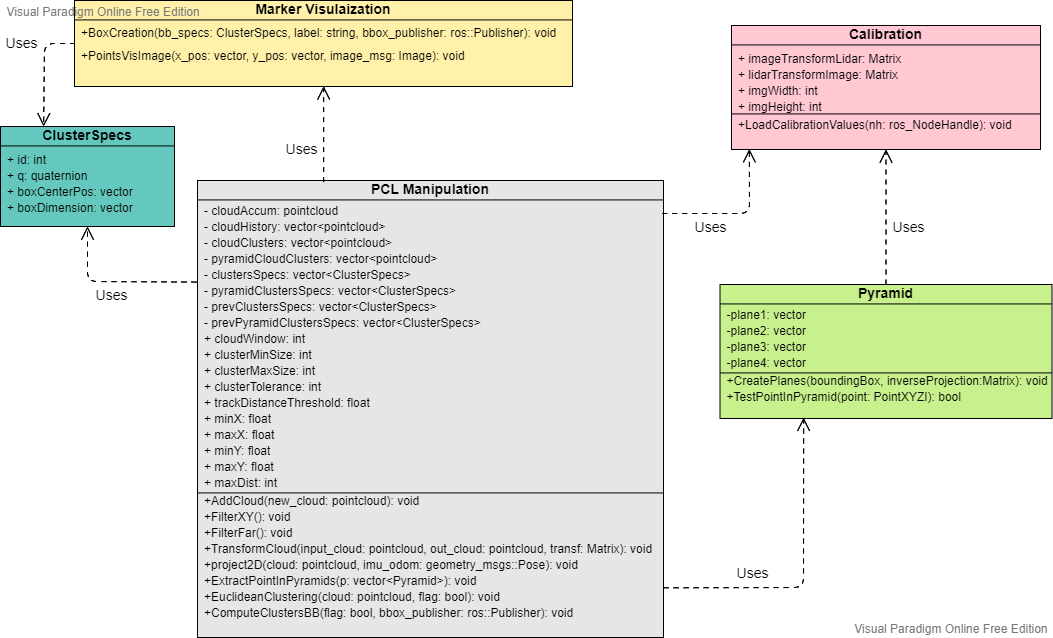} }
    \caption{System Class Diagram}%
    \label{fig:class_diag}%
\end{figure}

Another important diagram that represents the system from a "class" point of view is the Class Diagram depicted in figure ~\ref{fig:class_diag}. 

As indicated in the figure, the software has 5 main classes that are used to perform the operations described before in the component diagram. These classes are:

\begin{itemize}
    \item \textbf{Calibration:} This class handles the loading of the camera calibration parameters (image width and height, projection matrix, distortion coefficients, translation vector, etc). Then, using these parameters, it computes the transformation matrix from the image to lidar frame and vice versa.
    \item \textbf{Pyramid:} This is the class representing the pyramid in 3D corresponding to a 2D bounding box. The pyramid is represented by four planes where each plane is defined by its normal vector.
    \item \textbf{ClusterSpecs:} This class holds the specifications of a cluster representing an obstacle. The specifications include its tracking ID and its bounding box position, orientation, and dimensions. 
    \item \textbf{Marker Visualization:} This class holds the functions that deal with visualization of the markers on Rviz as well as visualization of the projected point cloud on the image plane (this is done for debugging purposes to ensure that the calibration parameters are accurate). It is subject to expansion in the future work.
    \item \textbf{PCL Manipulation:} The core class where all the operations on the point cloud and their attributes are defined. This class uses all the previously mentioned classes to do the various operations on the point clouds. 
    Among its attributes, it holds the variables of the accumulated cloud, accumulation window, and the vectors of cloud history, cloud clusters, and clusters extracted from pyramids. In addition, the clustering parameters are defined that determine the size of the clusters and their distance tolerance. 
\end{itemize}

\chapter{Experimental Analysis}
\label{chap:fourth}
\ifpdf
    \graphicspath{{Chapter4/Figures/PNG/}{Chapter4/Figures/PDF/}{Chapter4/Figures/}}
\else
    \graphicspath{{Chapter4/Figures/EPS/}{Chapter4/Figures/}}
\fi

In this chapter, we provide an analysis of the experiments conducted using both the original methodology and the software pipeline proposed in the previous chapter. The analysis includes both quantitative and qualitative evaluation on different datasets as well as an analysis of the execution time of the pipeline.

\section{Image-Based Obstacle Detection \& Tracking}

\subsection{Evaluation of Other Methods}
Given the public marine datasets that were detailed in the State of The Art chapter, a thorough scanning of the recent research papers in the field has been conducted to extract benchmarking results on these datasets. The scanning aimed at examining the evaluation results of different methodologies for obstacle detection in 2D images on the given datasets. These results were reported in a table~\ref{table_benchmark} in order to use it as a reference when we benchmark our method on these datasets. 

It is worth noting that a big disadvantage of most of the currently existing maritime datasets is that they do not provide a clear separation of training and testing sets. This led to inability to compare the evaluation results of different object detectors since every paper can select unknown set of images from the dataset and present the evaluation of their methodology based on that.

\begin{table}[ht]
\begin{center}
\caption{Benchmarking of some of the state of the art marine object detection algorithms on images}
\label{table_benchmark}
\resizebox{\linewidth}{!}{
\begin{tabular}{c c c c c c}
\hline
\textbf{Method} & \textbf{Dataset} & \textbf{test size} & \textbf{Performance} & \textbf{FPS}\\
\hline

Mask RCNN \cite{bovcon2021mods} & MODD2 & 81k & $F- score: 0.178$ & 9\\
YOLOv4 \cite{bovcon2021mods} & MODS & 81k & $F-score: 0.113$ & 25\\
Faster RCNN \cite{iancu2021aboships} & ABOships & 10k & $mAP: 0.35$ & -\\
EfficientDet \cite{iancu2021aboships} & ABOships & 10k & $mAP:0.34$ & -\\
SSM \cite{kristan2015fast} & MODD1 & 4.5k & $F-score$: 0.819 & 100\\
ISSM\_s \cite{bovcon2018stereo} & MODD2 & - & $F-score$: 0.778 & 11.1\\
SSM\_s \cite{bovcon2018stereo} & MODD2 & - & $F-score$: 0.356 & 90.13\\
EfficientDet\cite{nalamati2020automated} & SMD+SeaShips & 2731 & $mAP (IoU:0.5): 0.981$ & 13.00\\
CenterNet\cite{nalamati2020automated} & SMD+SeaShips & 2731 & $mAP (IoU:0.5): 0.893$ & 41.04\\
Faster RCNN\cite{shao2018seaships} & SeaShips & 32k & $mAP (IoU:0.5): 0.92$ & 7\\
YOLOv2\cite{shao2018seaships} & SeaShips & 32k & $mAP (IoU:0.5): 0.79$ & 91\\
modified YOLOv4\cite{liu2021sea} & SeaShips & 32k & $mAP (IoU:0.5): 0.95$ & 68\\
Mask RCNN\cite{moosbauer2019benchmark} & SMD (2 classes) & 40k & $F- score (IoU:0.3): 0.875$ & -\\

\hline

\end{tabular}
}
\end{center}
\end{table}

The main metric used by the community of object detection is the Average Precision (AP) measured usually at 0.5 threshold of intersection over union (IoU). Precision and Recall are generally defined as: 

$$Precision =\dfrac{TP}{TP + FP}$$

$$Recall =\dfrac{TP}{TP + FN}$$

In the field of object detection, the terms are defined as:
\begin{itemize}
    \item $TP$ (True Positives): predictions with $IoU \geq threshold$ with a ground truth box
    \item $FP$ (False Positives): predictions with $IoU < threshold$ with the ground truth boxes
    \item $FN$ (False Negatives): missing predictions of objects present in the ground truth.
\end{itemize}

In object detection tasks and challenges the term Average precision is referring to the Mean Average Precision (mAP) and reflects the precision averaged over all classes in the detection task. The precision is measured at 11 different recall points $(0, 0.1, 0,2, ... , 1)$ to estimate the area
under precision-recall curve. Formally:

$$AP = \dfrac{1}{11} \times (AP_{r}(0) + AP_{r}(0.1) + ... + AP_{r}(1.0))$$

The ground truth for this metric is the bounding boxes of the objects and their classes (depending on the classification of objects adopted in the datasets)

The $F- score$ is simply calculated given the precision and the recall as:

$$F- score = 2 \times \dfrac{Precision \times Recall}{Precision + Recall}$$

\subsection{Evaluation of Our Method}
\subsubsection{Quantitative Evaluation}
We conducted an evaluation of our previously described methodology for image-based obstacle detection and tracking on some of the mentioned public marine datasets. The datasets we chose are: SMD, MODD2, ABOships. We chose specifically these datasets since they are composed of video frames and they provide the ground truth files for the obstacles present in each frame. We couldn't evaluate our method on other datasets like Seaships because the images in this dataset are separate and they are not organized in the form of video frames. This doesn't comply with the required input format for our tracking module where it is necessary for the input to be successive frames drawn from a video.

We run YOLOv3 object detector on the video frames without thresholding the detections by confidence and by selecting only the Boat class as output. Then, we post-process the detections using the tracking module in online settings relying only on past frames. Using the updated detections after tracking, we evaluate them against the ground truth (by selecting only the classes similar to the Boat class and casting them to Boat) using an evaluation script provided by MS COCO. 

We report in table ~\ref{table_evaluation} the collective mAP (mean Average Precision) and mAR (mean Average Recall) obtained on all the videos in each of the three datasets and the IoU threshold used. The frame per second rate of the videos is also reported. Apparently, the obtained results for SMD and MODD2 are not high enough since the detector was not trained on these datasets. Therefore, their unique themes (environment, ships and boats shape, weather, etc) introduce a challenge to our detector. To improve the performance, a fine-tuning process of the detector can be conducted on part of these datasets. For ABOships, the obtained mAP and mAR were low because the FPS rate was extremely low (1 frame each 15 seconds). 

\begin{table}[h]
\begin{center}
\caption{Evaluation of Our Detection \& Tracking Method on Public Marine Datasets}
\label{table_evaluation}
\resizebox{\linewidth}{!}{
\begin{tabular}{c c c c c c c}
\hline
\textbf{Dataset} & \textbf{FPS} & \textbf{IoU threshold}& \textbf{mAP} & \textbf{mAR} & \textbf{Best mAP} \\
\hline

SMD\cite{SMD}   & not mentioned & 0.5 & 0.4 & 0.4 & 0.71 \\
MODD2\cite{bovcon2018stereo} & 10 & 0.15 & 0.44 & 0.48 & 0.644 \\
ABOships\cite{iancu2021aboships} & 0.067 & 0.15 & very low & very low & -- \\

\hline

\end{tabular}
}
\end{center}
\end{table}

\subsubsection{Qualitative Evaluation}

\begin{figure}%
    \centering
    \subfloat{{\includegraphics[width=11cm, height=8cm]{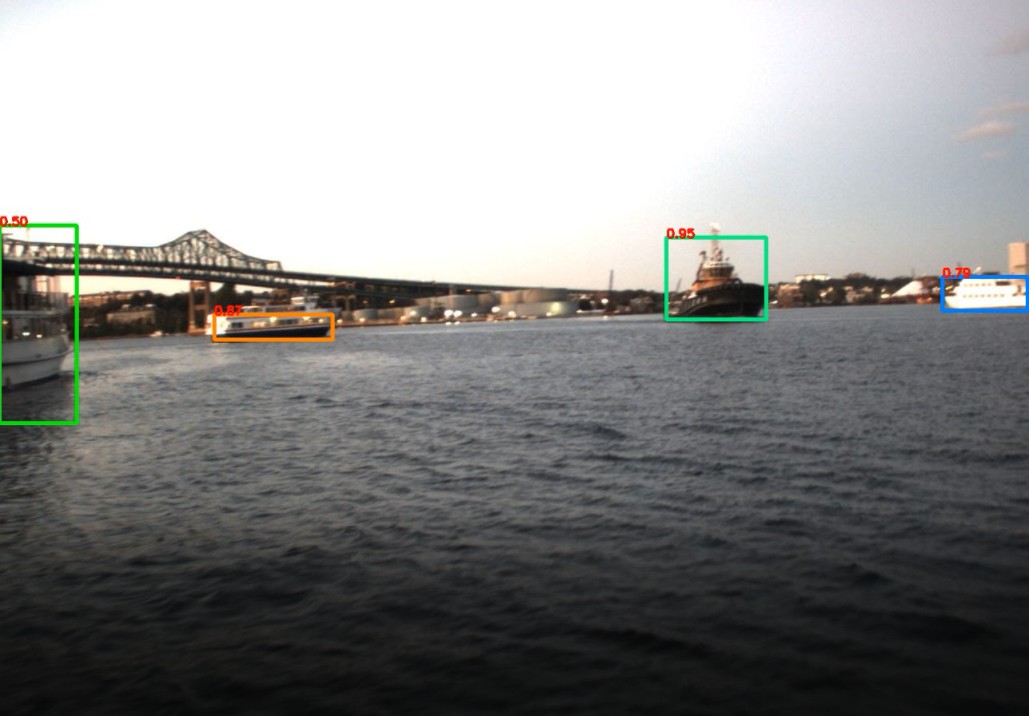} }}%
    \qquad
    \subfloat{{\includegraphics[width=11cm, height=8cm]{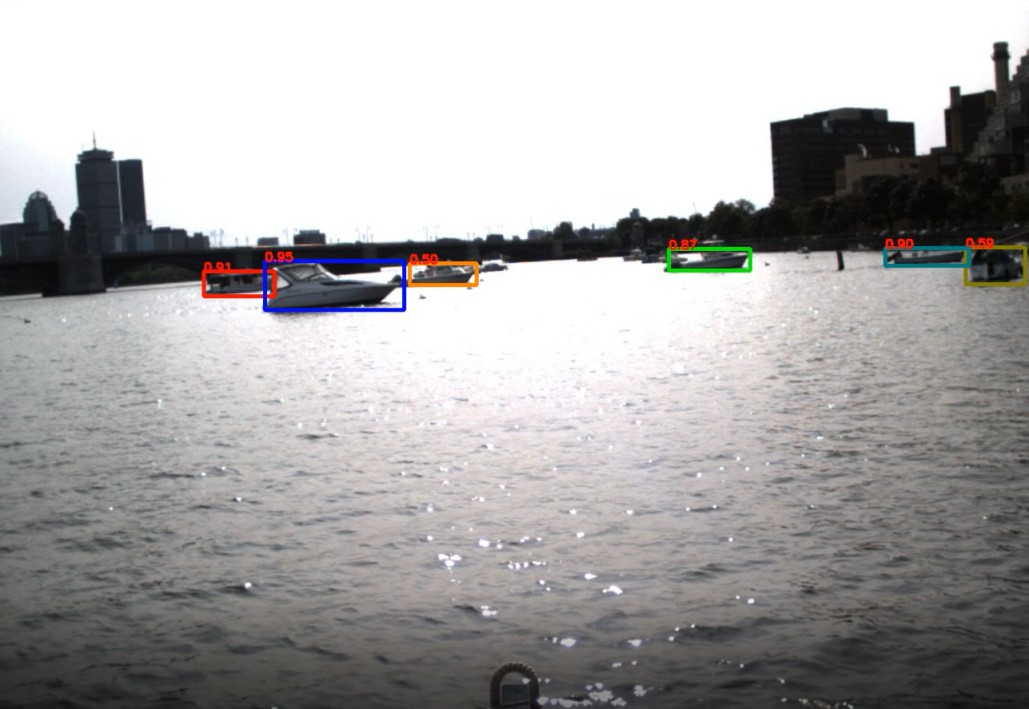} }}%
    \caption{Good Detection Examples}%
    \label{fig:det_g}%
\end{figure}

\begin{figure}%
    \centering
    {\includegraphics[width=11cm, height=8cm]{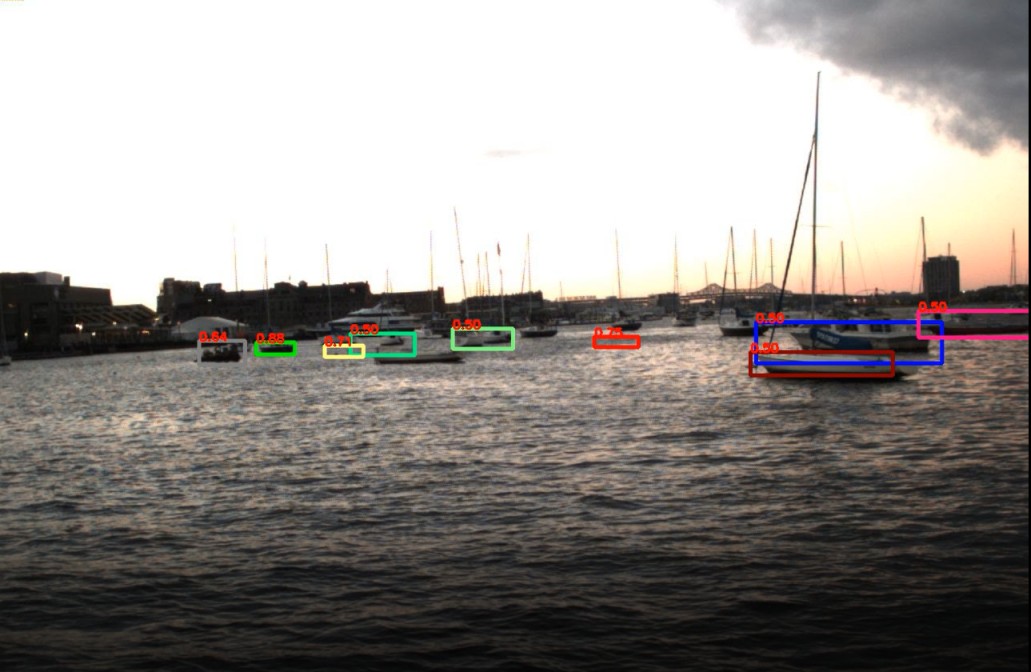} }
    \caption{Detection Problem in Cluttered Environments}%
    \label{fig:det_p_cluttered}%
\end{figure}

\begin{figure}%
    \centering
    {\includegraphics[width=11cm, height=8cm]{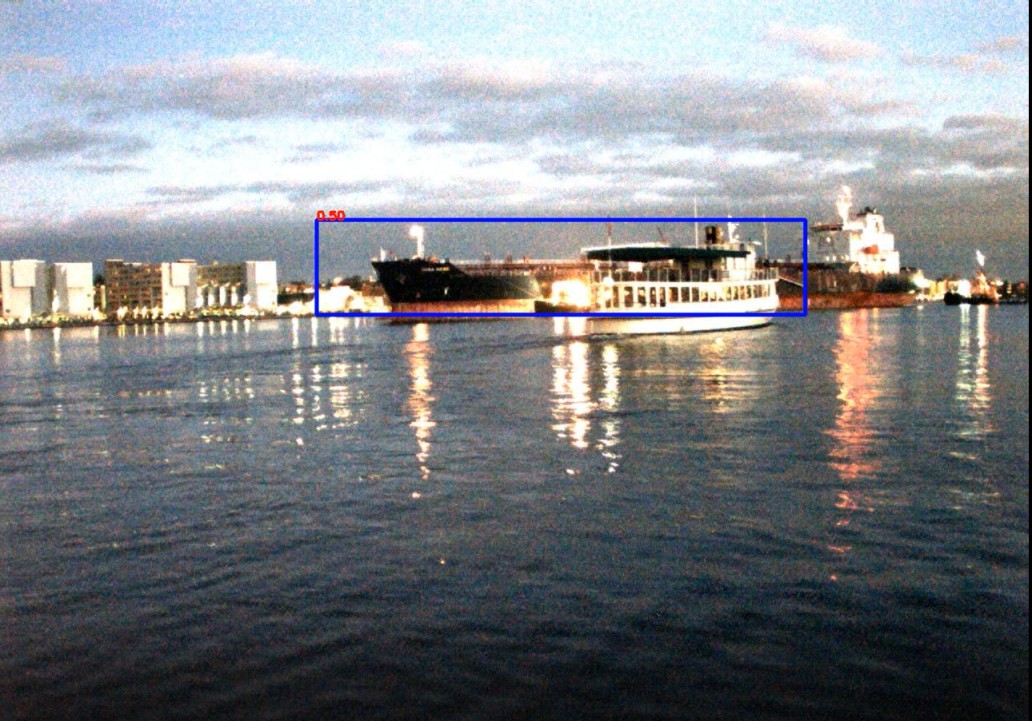} }
    \caption{Detection Problem of Overlapping Obstacles}%
    \label{fig:det_p_overlap}%
\end{figure}

\begin{figure}%
    \centering
    {\includegraphics[width=11cm, height=8cm]{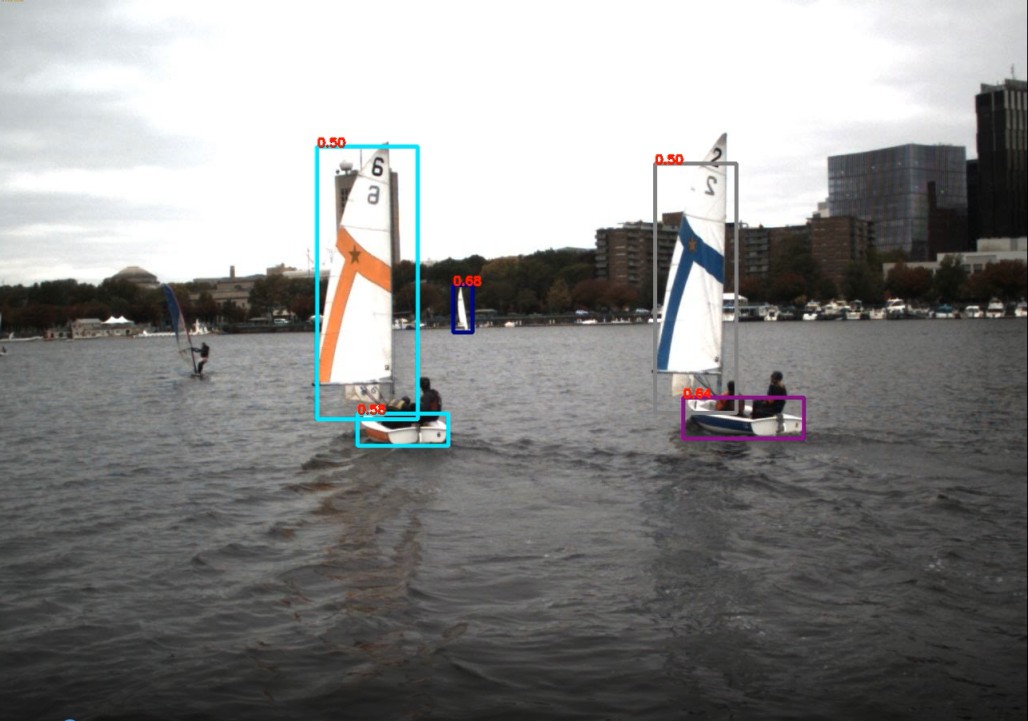} }
    \caption{Detection Problem of Sailboats}%
    \label{fig:det_p_sailboat}%
\end{figure}

We run the detection \& tracking pipeline on some of the videos of the MIT dataset and evaluate the performance qualitatively since there are no ground truth available for this dataset. The following are the most important observations:

\begin{itemize}
    \item The system performance is overall excellent with large and relatively close obstacles, see figure~\ref{fig:det_g}.
    \item The detections of far away obstacles get lost at some times leading them to getting assigned new track IDs
	\item The performance of the detector is greatly declined in cluttered environments, see figure~\ref{fig:det_p_cluttered}.
	\item Overlapping obstacles get detected as one big obstacle, see figure~\ref{fig:det_p_overlap}.
	\item Tracks when the USV is moving up and down (waves) get lost in the middle leading to identity confusion
	\item Sailboats sometimes are detected as two objects, see figure~\ref{fig:det_p_sailboat}.
\end{itemize}

Most of the mentioned challenges can be overcome by manually annotating some of the video frames in the MIT dataset and fine-tune the YOLO detector on them. This can be done in the future work. 

\section{Evaluation of The Whole Pipeline}
By adopting the software architecture that was presented in chapter 4, we qualitatively evaluate its performance on the MIT dataset. To our knowledge, the MIT dataset is the only available public marine dataset that contains synchronized Lidar and camera data. We choose different scenarios and conditions in the dataset for testing our system and analyze the obtained results. 

In these experiments we choose the parameters as following:
\begin{itemize}
    \item \textbf{cloud accumulation window:} 3 successive clouds
    \item \textbf{cluster Min Size:} 30 points
    \item \textbf{cluster Max Size:} 10000 points
    \item \textbf{cluster Tolerance:} 3.0 meters
    \item \textbf{track Distance Threshold:} 5.0 meters
\end{itemize}

The LiDAR and camera data are received through the ROS topics with a frequency of ~10 Hz, and YOLO detector can process an average of 25 FPS (faster than the reception rate of the images).

\subsection{MIT Dataset Processing}
\subsubsection{Metadata}
\begin{figure}%
    \centering
    {\includegraphics[width=12cm, height=8cm]{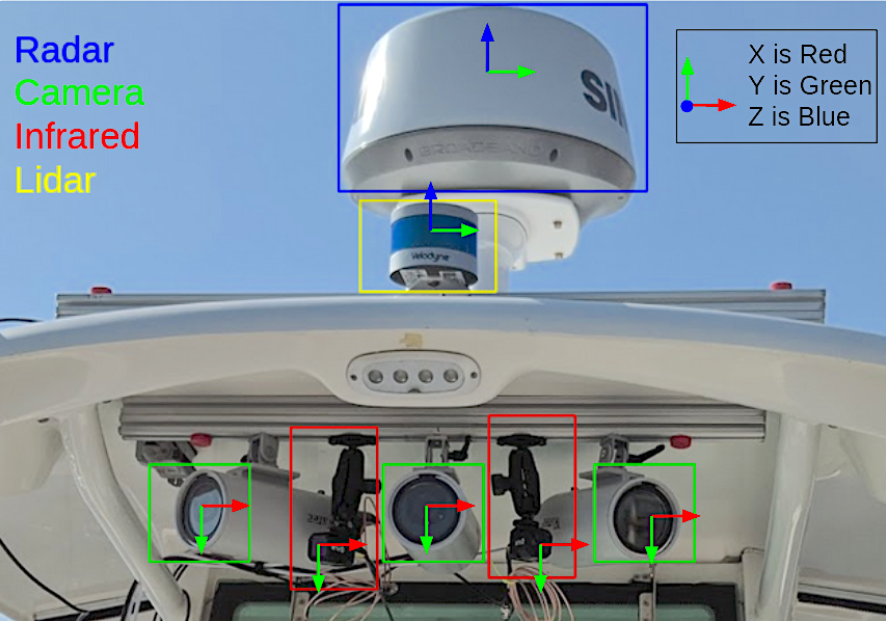} }
    \caption{Sensors Frames in the MIT Vehicle}%
    \label{fig:frames}%
\end{figure}

To correctly use the MIT dataset for testing our system, we had to perform some investigations on the provided metadata. The dataset is composed of 35 different datasets where each of them contains the ROS bag file of the collected data as well as the transformations of sensors coordinate frames relative to vehicle body frame, see figure~\ref{fig:frames}. In addition, the three cameras (right, center, left) calibration files are provided that include the projection matrix from the 3D camera frame to the 2D image frame.

\begin{figure}%
    \centering
    {\includegraphics[width=11cm, height=8cm]{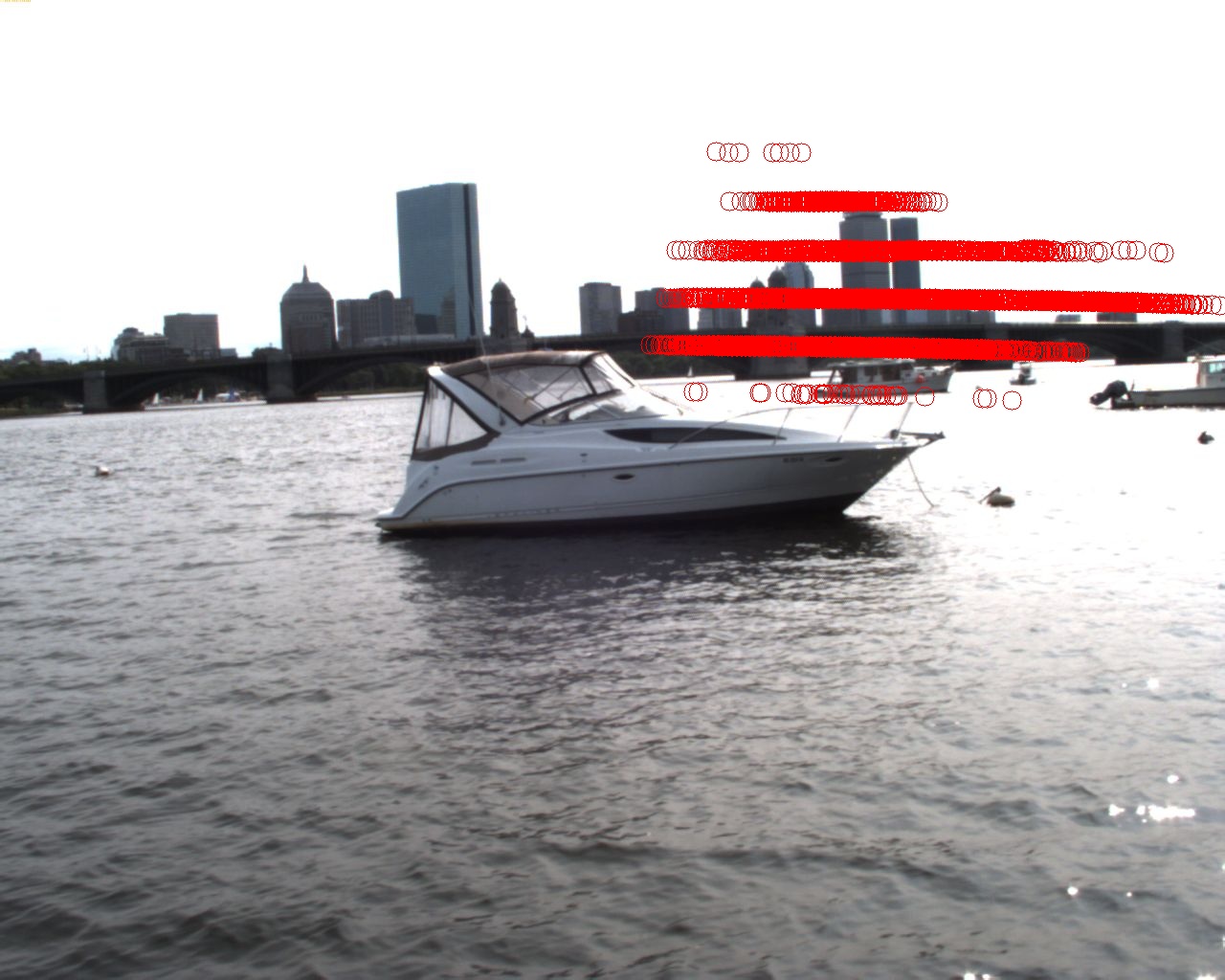} }
    \caption{Projection of point cloud on image using provided metadata}%
    \label{fig:calib_prob}%
\end{figure}

\begin{figure}%
    \centering
    {\includegraphics[width=11cm, height=8cm]{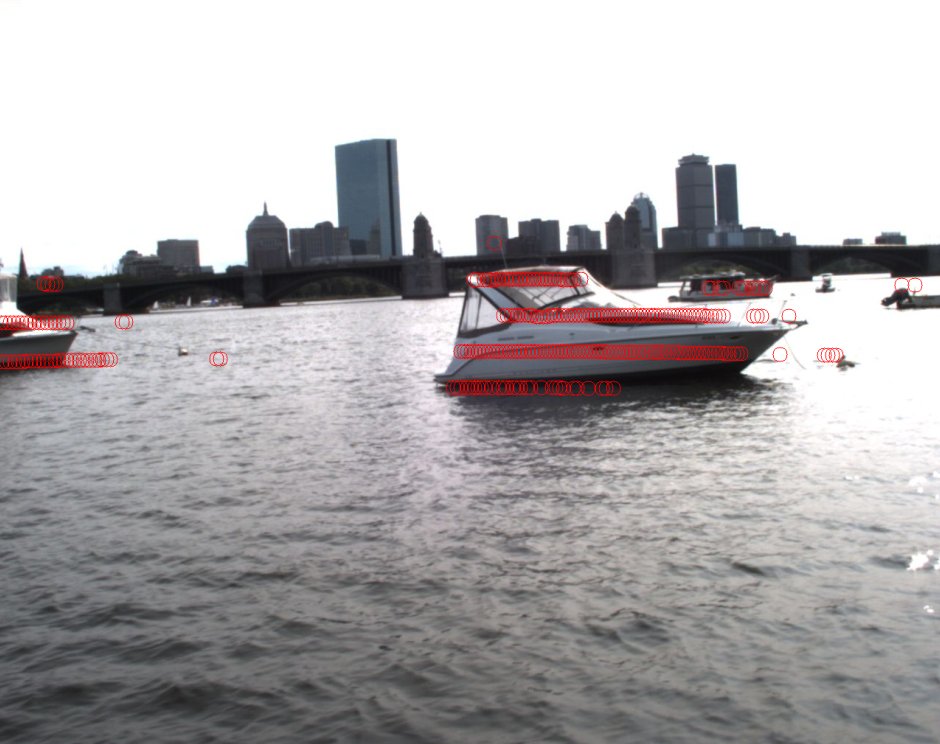} }
    \caption{Projection of point cloud on image after tuning the metadata}%
    \label{fig:calib_corr}%
\end{figure}

To test the accuracy of the provided metadata, we implement a simple module that projects the Lidar point cloud on its synchronized image frame in real time and we investigate the quality of the projection on the image. After investigation, we found that the given metadata needs some tuning, see figure~\ref{fig:calib_prob}. After tuning the given transformations, the projection becomes accurate, see figure~\ref{fig:calib_corr}.

\subsubsection{IMU Data}

We also investigated the IMU data published in the ROS bag files, and we found that they are noisy and cannot be used as a reference for projecting the point cloud on the sea plane. For this reason, we opt to assuming that the orientation of the USV frame doesn't change with respect to the reference frame of the sea plane and we apply the projection by simply zeroing out the Z-component of the point cloud.

\subsection{Performance in Near-Port Environments}

We run the system on a dataset corresponding to a near-port scenario and choose the data coming from the \textbf{left} camera as an input. We visualize in real-time the output bounding boxes on Rviz that are extracted from both the pyramids-cloud-clustering approach and the all-cloud-clustering approach. 

\begin{figure}%
    \centering
    \subfloat{{\includegraphics[width=14cm, height=4.8cm]{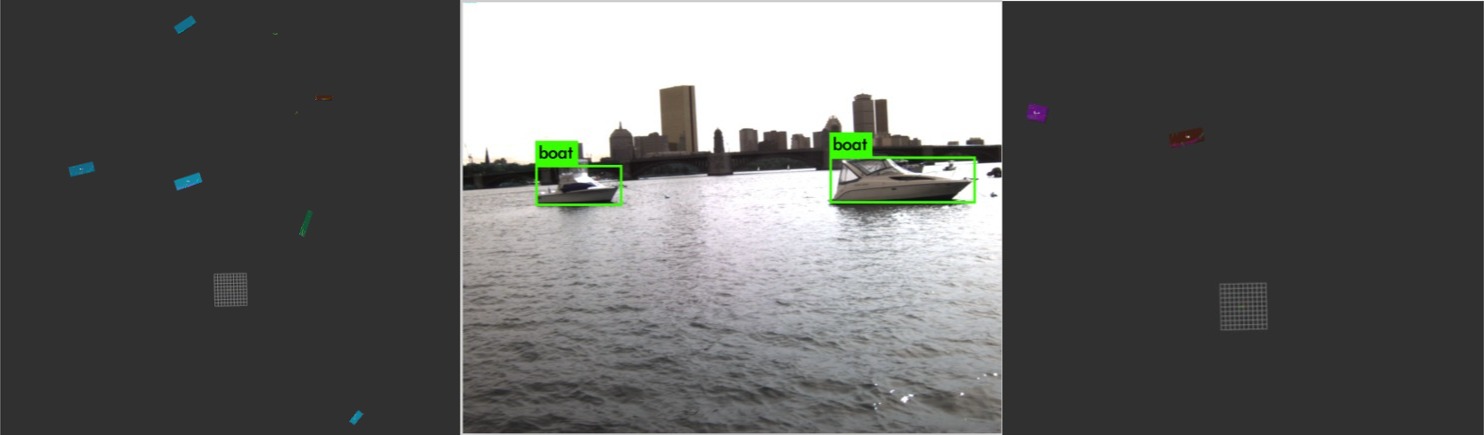} }}%
    \qquad
    \subfloat{{\includegraphics[width=14cm, height=4.8cm]{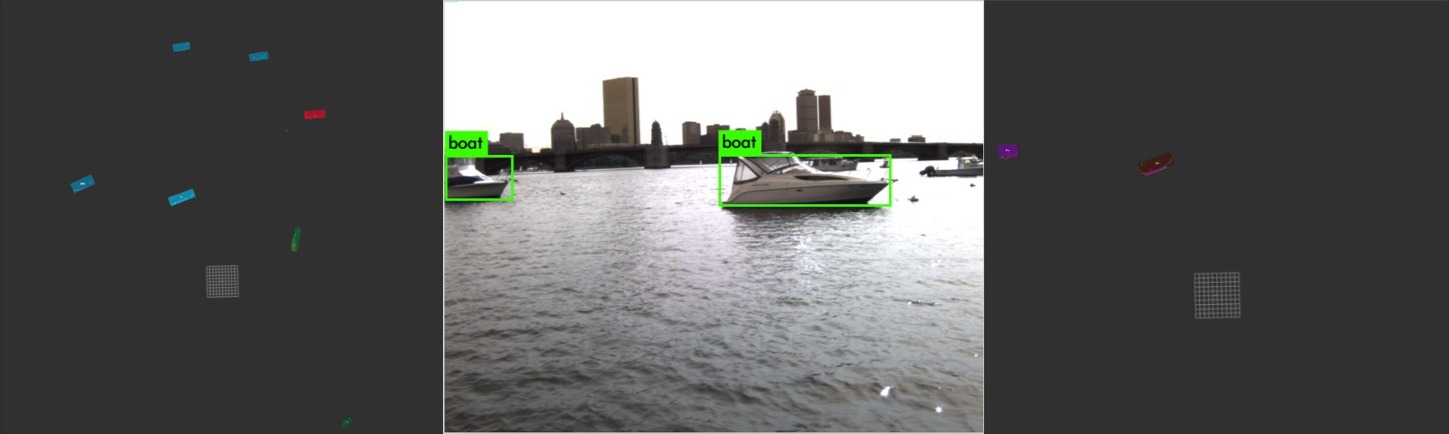} }}%
    \qquad
    \subfloat{{\includegraphics[width=14cm, height=4.8cm]{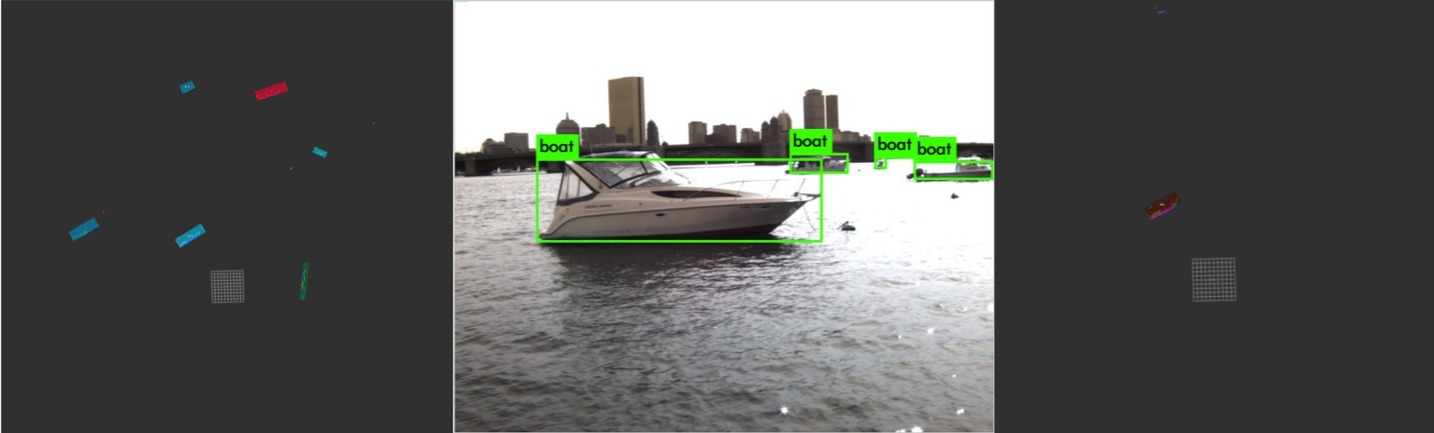} }}%
    \caption{Near-Port Results 1 of all-cloud-clustering approach (left) and pyramids-cloud-clustering approach (right). Bird's-eye view is used}%
    \label{fig:det_p1}%
\end{figure}

\begin{figure}%
    \centering
    \subfloat{{\includegraphics[width=14cm, height=4.5cm]{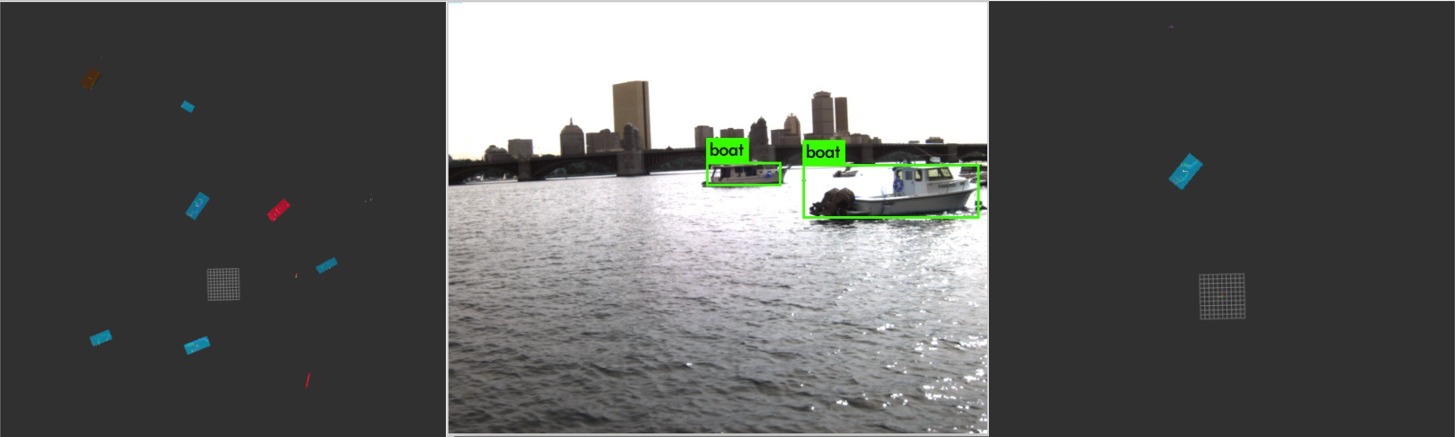} }}%
    \qquad
    \subfloat{{\includegraphics[width=14cm, height=4.5cm]{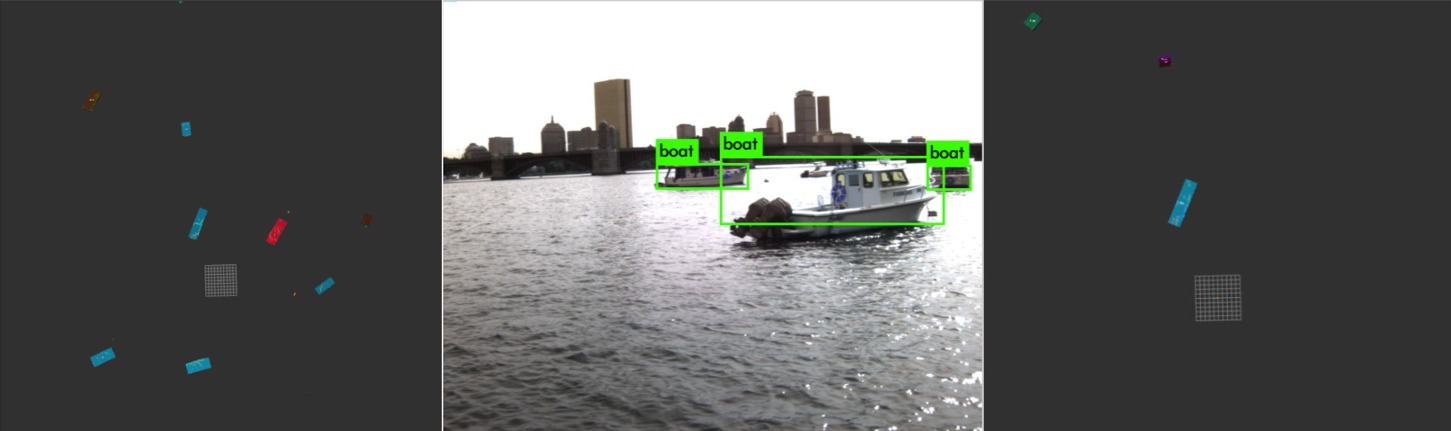} }}%
    \qquad
    \subfloat{{\includegraphics[width=14cm, height=4.5cm]{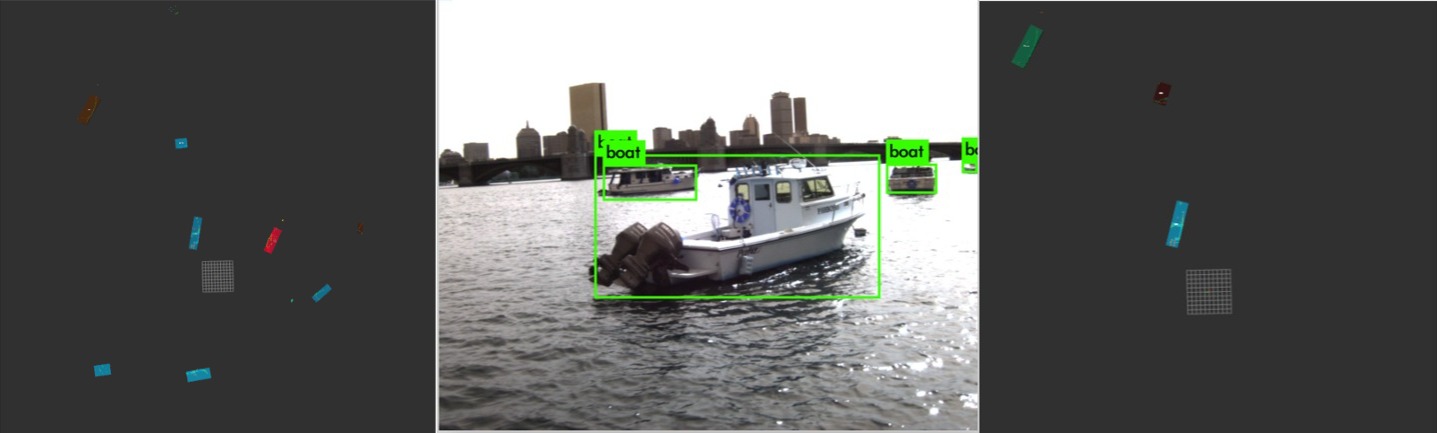} }}%
    \qquad
    \subfloat{{\includegraphics[width=14cm, height=4.5cm]{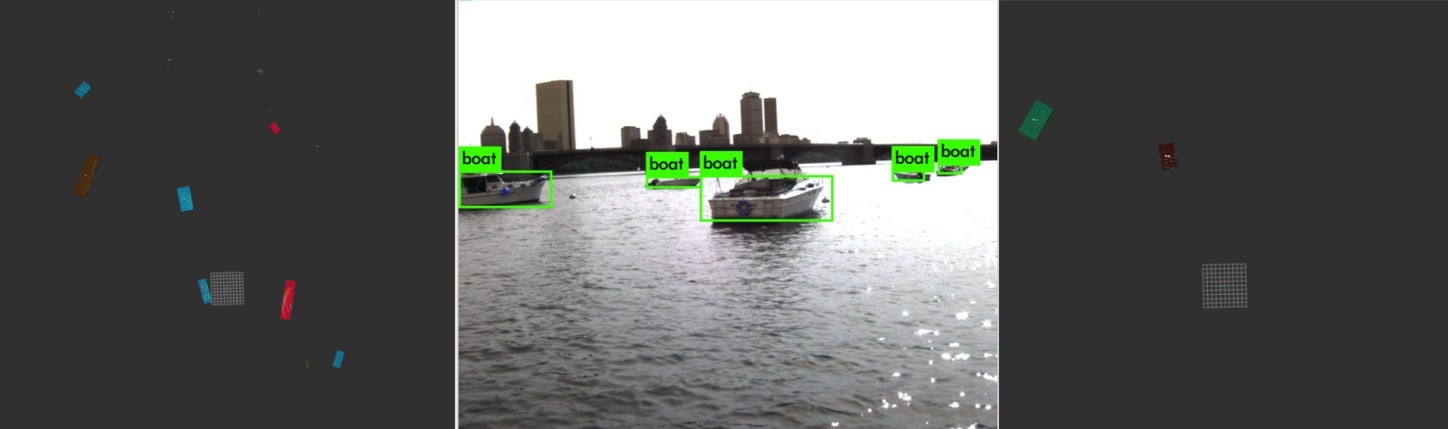} }}%
    \caption{Near-Port Results 2 of all-cloud-clustering approach (left) and pyramids-cloud-clustering approach (right). Bird's-eye view is used}%
    \label{fig:det_p2}%
\end{figure}

In figure~\ref{fig:det_p1} and figure~\ref{fig:det_p2} the bounding boxes of the clusters extracted using both approaches are presented at successive time steps along with the corresponding output from YOLO. Each box colour represents its tracking ID (some colours may look the same but they are slightly different in the shade). The center of the white grid in the middle represents the USV frame.

The following observations can be marked:
\begin{itemize}
    \item The pyramids-cloud-clustering approach works well with big objects. However, for far objects, their YOLO bounding box is a bit tight and, hence, its corresponding pyramid may not include any points in the point cloud, see the first and last frames of figure~\ref{fig:det_p2}.
    \item The above problem is overcome in the all-cloud-clustering approach where the whole point cloud is considered for clustering, see the first and last frames of figure~\ref{fig:det_p2}.
    \item Accumulating the point cloud (accumulation window = 3) overcomes the sparsity of a single point cloud and greatly improves the clustering output.
    \item The tracking by distance method works quite well and is able to identify most of the obstacles over time. However, it is not accurate all the time.
\end{itemize}

\subsection{Performance in Cluttered Environments}

\begin{figure}%
    \centering
    \subfloat{{\includegraphics[width=14cm, height=4.5cm]{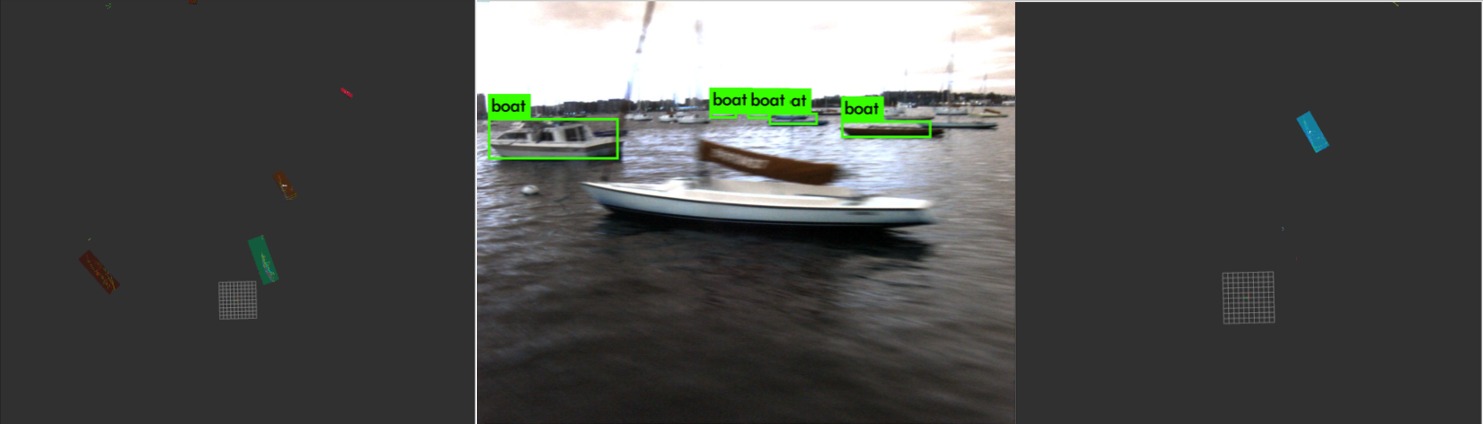} }}%
    \qquad
    \subfloat{{\includegraphics[width=14cm, height=4.5cm]{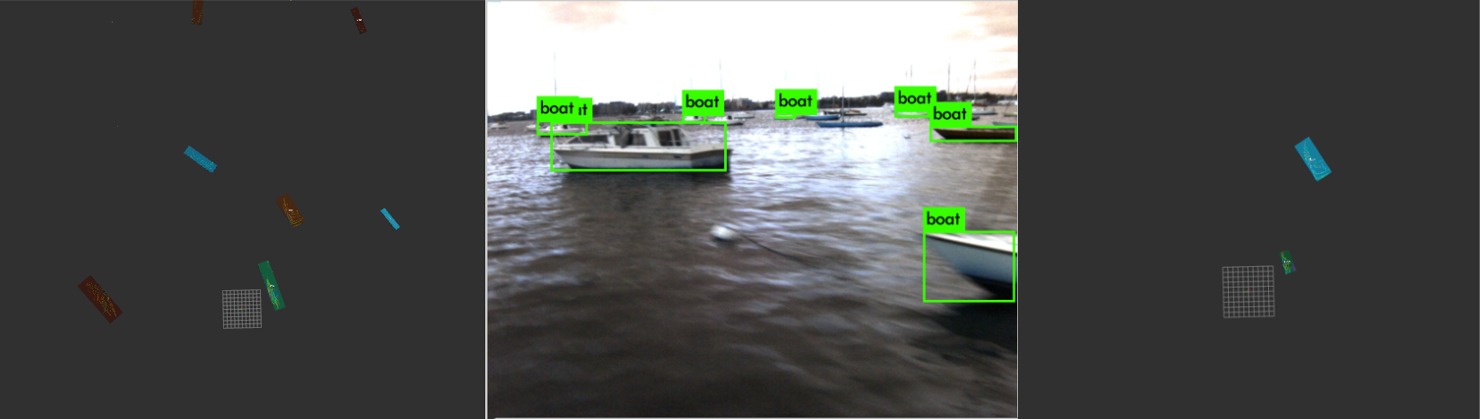} }}%
    \qquad
    \subfloat{{\includegraphics[width=14cm, height=4.5cm]{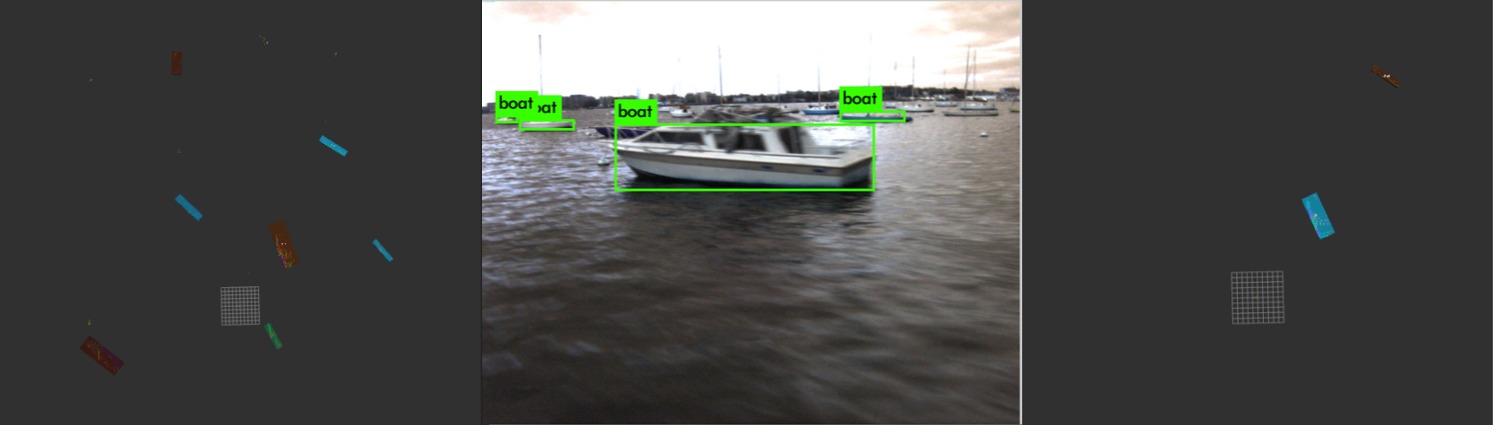} }}%
    \qquad
    \subfloat{{\includegraphics[width=14cm, height=4.5cm]{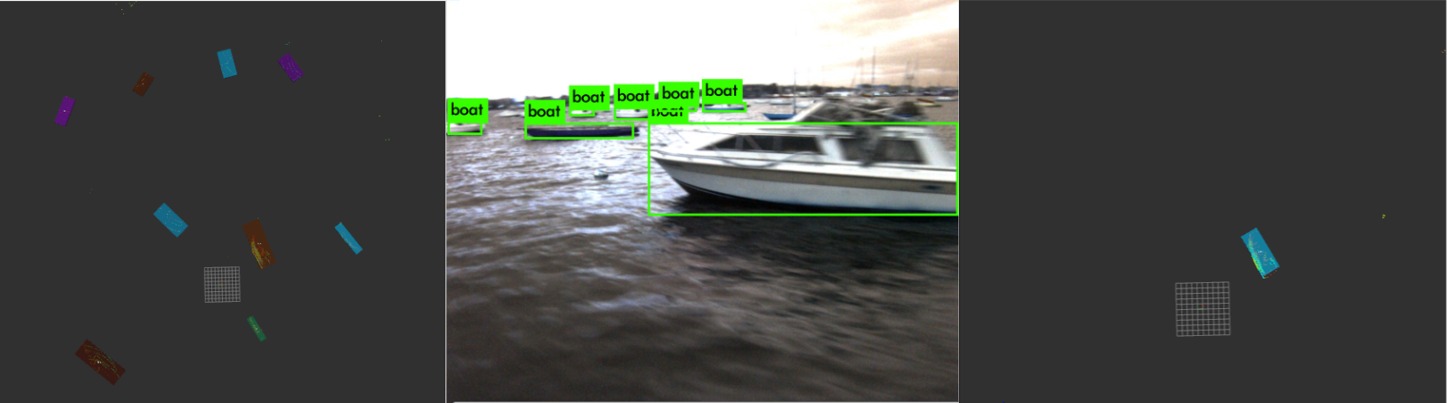} }}%
    \caption{Cluttered Environment Results of all-cloud-clustering approach (left) and pyramids-cloud-clustering approach (right). Bird's-eye view is used}%
    \label{fig:det_clut}%
\end{figure}

The system was tested on a dataset corresponding to a cluttered scenario with choosing the data coming from the \textbf{right} camera as an input. The results are presented in figure~\ref{fig:det_clut}.

The following observations can be marked:
\begin{itemize}
    \item The multiple far objects caused the failure of the pyramids-cloud-clustering approach to detect them.
    \item When YOLO misses the detection of an object, it doesn't get detected using the pyramids-cloud-clustering approach. On the other hand, this is not an issue for the the all-cloud-clustering approach. See the first frame of figure~\ref{fig:det_clut}
    \item The output of the all-cloud-clustering approach is much more reliable in this scenario than the pyramids-cloud-clustering approach.
    \item The tracking by distance method works quite well and is able to identify most of the obstacles over time. However, it is not accurate all the time.
\end{itemize}

\subsection{Performance in Open Sea Environments}

\begin{figure}%
    \centering
    \subfloat{{\includegraphics[width=14cm, height=4.5cm]{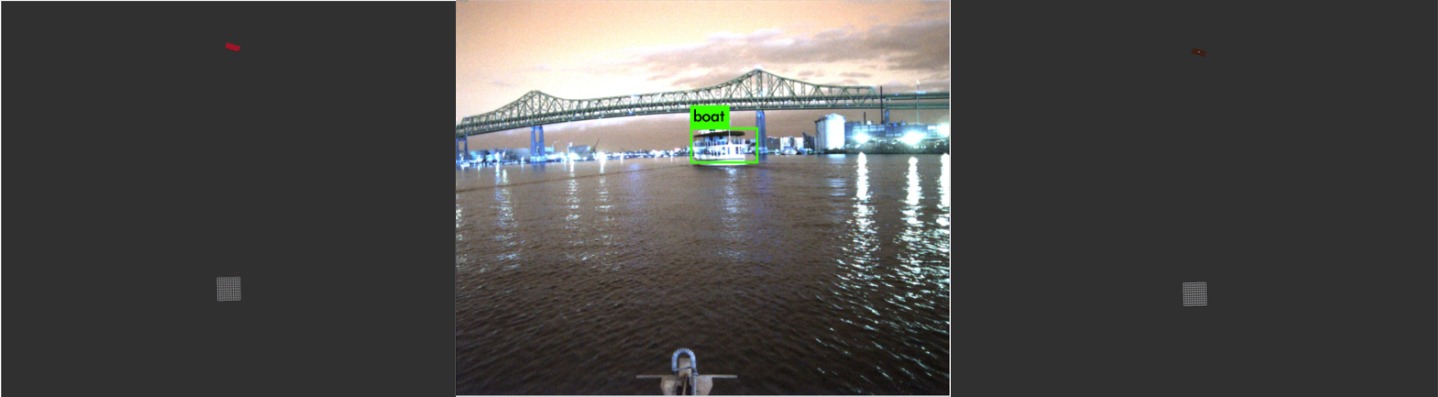} }}%
    \qquad
    \subfloat{{\includegraphics[width=14cm, height=4.5cm]{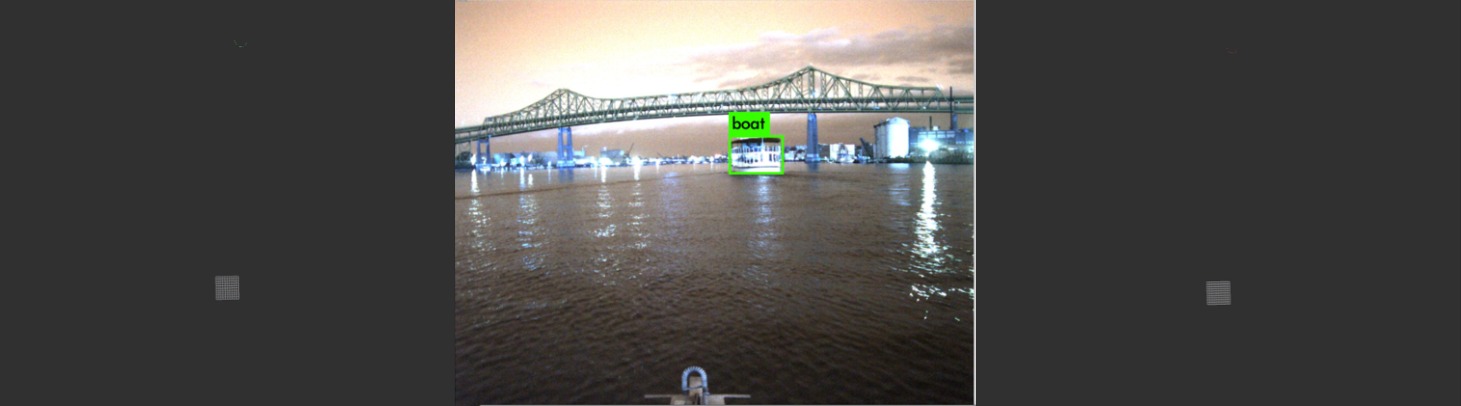} }}%
    \qquad
    \subfloat{{\includegraphics[width=14cm, height=4.5cm]{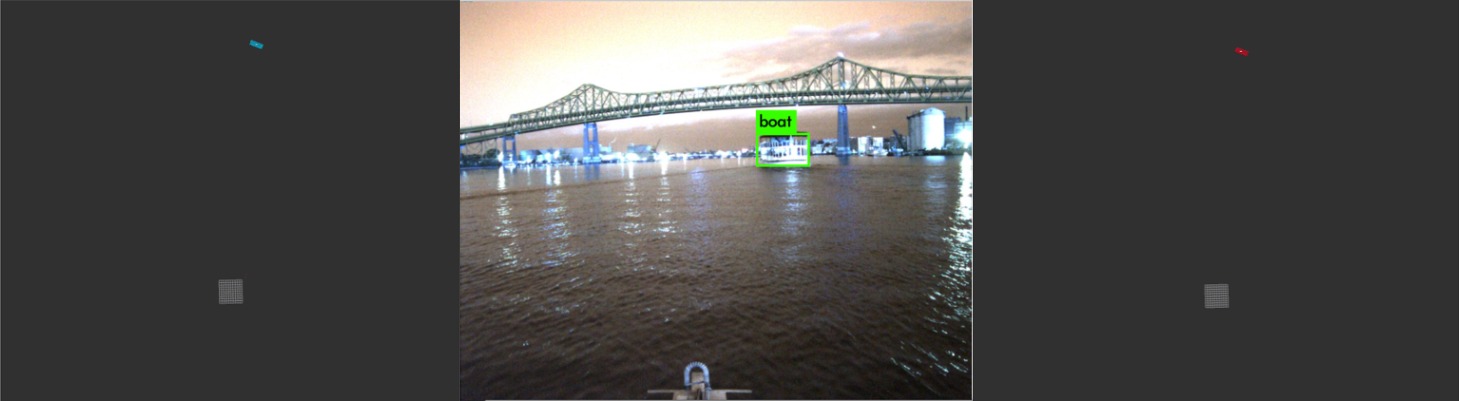} }}%
    \qquad
    \subfloat{{\includegraphics[width=14cm, height=4.5cm]{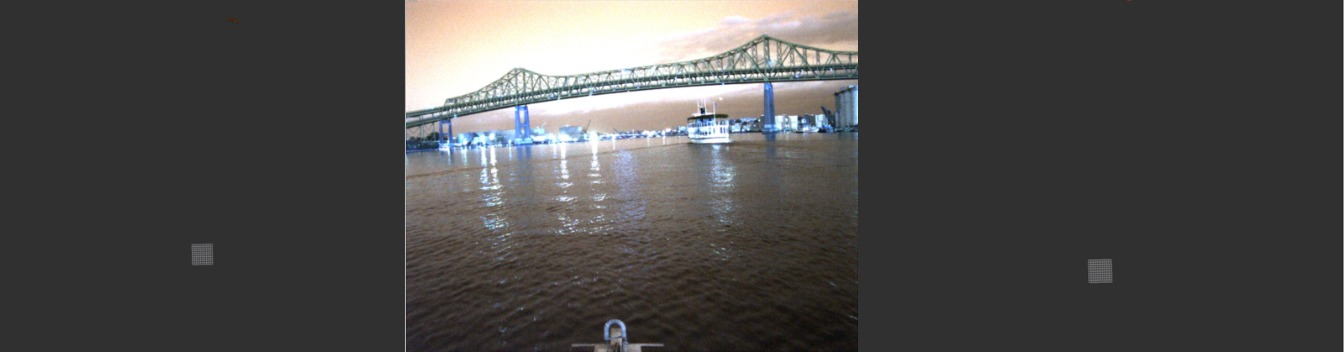} }}%
    \caption{Open Sea Environment Results of all-cloud-clustering approach (left) and pyramids-cloud-clustering approach (right). Bird's-eye view is used}%
    \label{fig:det_far}%
\end{figure}

The system was tested on a dataset corresponding to a far-from-port scenario with choosing the data coming from the \textbf{center} camera as an input. The results are presented in figure~\ref{fig:det_far}.

\begin{figure}%
    \centering
    {\includegraphics[width=14cm, height=6.7cm]{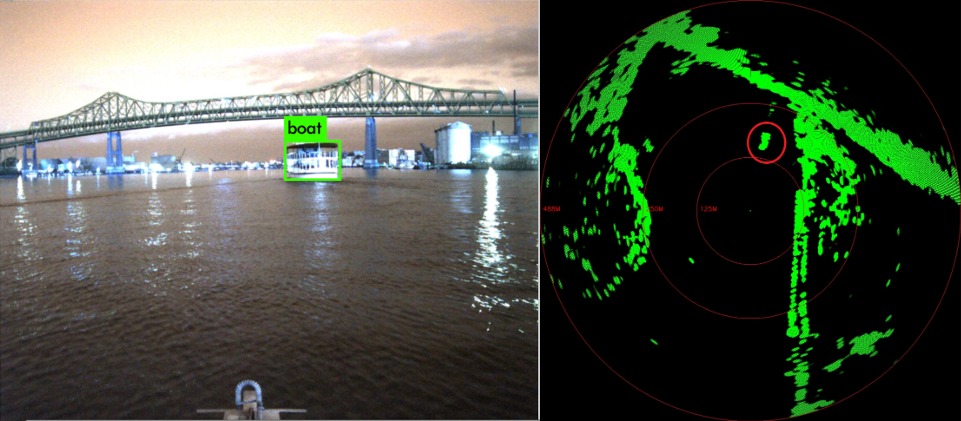} }
    \caption{Far obstacles appearing in the RADAR scan}%
    \label{fig:rad}%
\end{figure}

In such scenario, both approaches perform badly. The main reason for that is the few lidar points that reflect from relatively far objects. This makes it difficult for the clustering algorithm to identify these small number of points as a cluster (minimum cluster size is set to 30 points), and they are regarded as noise. In such scenarios, using RADAR in the place of LiDAR can be more promising because far objects appear in the RADAR data clearly, see figure~\ref{fig:rad}.

\subsection{Performance in Snowy Environments}

\begin{figure}%
    \centering
    {\includegraphics[width=14cm, height=6.7cm]{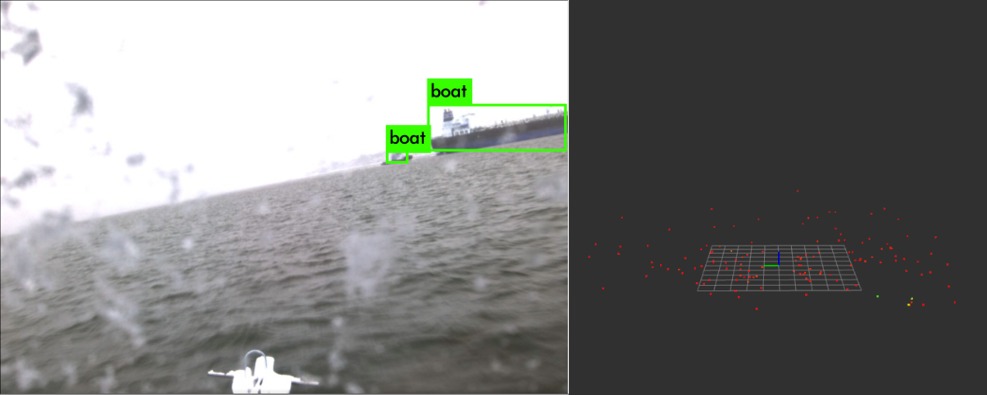} }
    \caption{Snowy Environment Result}%
    \label{fig:snow}%
\end{figure}

Our system was also tested on datasets corresponding to some extreme weather conditions like snow, see figure~\ref{fig:snow}. In this case, it was found that the received point cloud was immensely noisy due to the fact that snow flakes reflect the LiDAR laser beams. In such condition, our detection and tracking methodology cannot be useful.

\subsection{Execution Time Analysis}

\begin{figure}%
    \centering
    \subfloat{{\includegraphics[width=10cm, height=4cm]{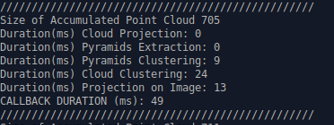} }}%
    \qquad
    \subfloat{{\includegraphics[width=10cm, height=4cm]{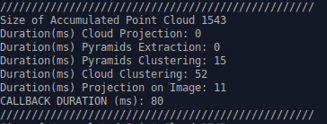} }}%
    \qquad
    \subfloat{{\includegraphics[width=10cm, height=4cm]{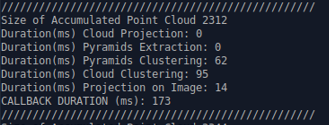} }}%
    \qquad
    \subfloat{{\includegraphics[width=10cm, height=4cm]{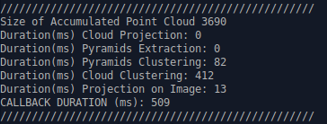} }}%
    \caption{Execution time computed for the main blocks of the system and for the whole pipeline (CALLBACK DURATION)}%
    \label{fig:time}%
\end{figure}

Since execution time is crucial in marine navigation missions where real time operation is required. An analysis has been performed on the computation time of the main blocks of the system as well as the whole pipeline. In figure~\ref{fig:time}, the calculated execution times are demonstrated in accordance with the size of the accumulated point cloud. 

It can be clearly noticed that execution times of the pyramids-cloud-clustering and all-cloud-clustering approaches have direct proportion with the point cloud size. This is because the clustering algorithm works in both cases and the computation time of such algorithm depends on the cloud size. On the other hand, cloud projection on the sea plane, pyramids extraction from YOLO bounding boxes, and cloud projection on image have negligible execution times and are not affected by the cloud size.

\begin{figure}%
    \centering
    {\includegraphics[width=10cm, height=5cm]{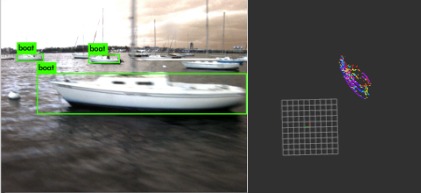} }
    \caption{The lag caused by long execution time}%
    \label{fig:lag}%
\end{figure}

As mentioned before, the input lidar and camera data are received with a frequency of ~10Hz. This means that an execution time of maximum 100 ms for the whole pipeline is tolerable and will not lead to the loss of any input data. However, in the third and fourth samples in figure~\ref{fig:time} where the size of the point cloud is greater than 2000 points, the pipeline duration exceeds 100 ms. This leads to loss of input lidar data and causes a lag effect on the accumulated cloud that can be viewed in figure~\ref{fig:lag}. 

From what was demonstrated we can deduce the following: 

\begin{itemize}
    \item Despite the fact that the all-cloud-clustering approach gives a full map of the surrounding obstacles, it has a high computation time that cannot be tolerated in very cluttered environments.
    \item On the contrary, the pyramids-cloud-clustering approach has low execution time as it shrinks the size of the processed cloud extracting only the points corresponding to obstacles in the camera view, however, it gives a partial map of the surrounding obstacles (only the obstacles lying in the camera view). 
\end{itemize}

\subsection{Hybrid Approach}

\begin{figure}%
    \centering
    {\includegraphics[width=14cm, height=6.7cm]{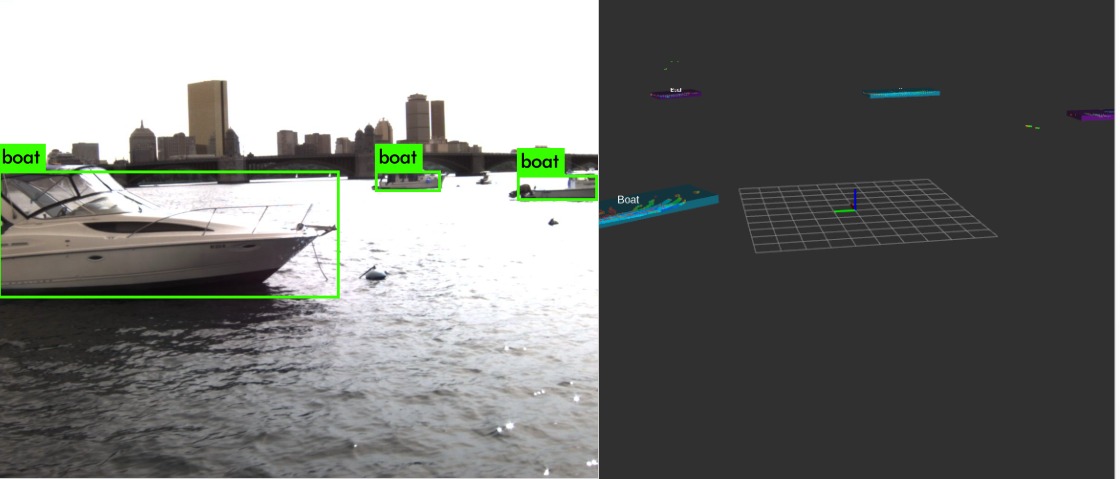} }
    \caption{Hybrid Approach Result}%
    \label{fig:hyb}%
\end{figure}

One of the main advantages of using the camera as a sensor is that it can provide semantic information about the obstacles (most importantly, the obstacle class). For this reason, we conducted an experiment where we use the all-cloud-clustering approach for building the obstacles map and we use the pyramids approach only to give semantic classes to the matching obstacles in the map. In such case, the obstacles map will cover all the surroundings, and only obstacles appearing in the camera view will have labels indicating their class while the rest of obstacles will be unknown, see figure~\ref{fig:hyb}.

\def\baselinestretch{1}
\chapter{Conclusion \& Future Improvements}
\label{chap:conclusions}
\ifpdf
    \graphicspath{{Conclusions/Figures/PNG/}{Conclusions/Figures/PDF/}{Conclusions/Figures/}}
\else
    \graphicspath{{Conclusions/Figures/EPS/}{Conclusions/Figures/}}
\fi
\def\baselinestretch{1.66}

Developing a robust and effective obstacle detection and tracking system for USVs at marine environments is not a straight-forward task. Research efforts have been made in this area during the past years by GRAAL lab at the university of Genova that resulted in a methodology for detecting and tracking obstacles on the image plane and, then, locating them in the 3D LiDAR point cloud. In this work, we continue on the developed system by firstly evaluating its performance on recently published marine datasets and we integrate the different blocks of the system on ROS platform where we could conduct real time testing on synchronized LiDAR and camera data collected in various marine conditions available in the MIT marine datasets in the form of ROS bag files. 

We presented a thorough experimental analysis of the results obtained using two approaches; one that uses sensor fusion between the camera and LiDAR to detect and track the obstacles and the other uses only the LiDAR point cloud for the detection and tracking. In the end, we provided a hybrid approach that merges the advantages of both approaches to build an informative obstacles map of the surrounding environment.

Developing and improving the complete system in the future can be achieved through successive steps of research approaches:
\begin{enumerate}
    \item The obstacle detector in the images coming from the camera can be improved by investigating more recent object detection neural networks like Faster RCNN~\cite{ren2015faster}, YOLOv4~\cite{bochkovskiy2020yolov4}, YOLOv5~\cite{yolov5}, or the most recent and fastest YOLOv6~\cite{yolov6}. It is more promising to try fine-tuning such models on marine datasets like MODD2~\cite{bovcon2018stereo}, SMD~\cite{SMD}, and Seaships~\cite{shao2018seaships} where various classes of obstacles are annotated. Also, inference time of these networks shall be computed and taken into consideration given the hardware computation capabilities. 
    
    \item Investigating the possibility of integrating data coming from another sensor like Infrared (IR) Camera and RADAR to boost the accuracy and robustness in various conditions. IR cameras are especially useful at night times and RADAR is useful for detecting far obstacles. Detection neural networks can also be fine-tuned on the publicly available annotated IR datasets.
    
    \item Working on the vehicle navigation including path planning (both global and local) and control after producing the 3D scene map including the obstacles bounding boxes and their trajectories. Different path planning approaches can be discussed and tested like graph-based search algorithms (Dijkstra, A*, etc), sampling-based algorithms (PRM, RRT and its variants, etc) that do not provide optimal solutions yet they are fast especially in complex environments, or learning-based algorithms (neural networks, reinforcement learning).

    \item The implementation and testing of the software system on a hardware platform is one of the main tasks that shall be tackled. This can be done after reaching a solid and robust milestone of the software system. 
    
    \item Testing the overall system in open sea scenarios ranging from simple to corner cases and analyze the performance of the system and its weak points. During these tests, rosbag files can be used to record data coming from the sensors and the data being published by the system in order to later use them in the analysis of errors and for code improvement. 

\end{enumerate}


\bibliographystyle{ieeetr}
\renewcommand{\bibname}{References}           
\bibliography{references}          
\nocite{*}

\end{document}